\documentclass[runningheads]{llncs}

\usepackage{eccv}

\usepackage{eccvabbrv}
\usepackage{amsmath}

\usepackage{graphicx}
\usepackage{booktabs}
\usepackage{tabularx}
\usepackage[accsupp]{axessibility}  
\usepackage{makecell}
\usepackage{array}
\usepackage{nicematrix}
\usepackage{array}

\RequirePackage[breakable,skins]{tcolorbox}
\RequirePackage{xcolor}

\newcolumntype{C}[1]{>{\centering\arraybackslash}m{#1}}

\usepackage[pagebackref,breaklinks,colorlinks]{hyperref}

\usepackage{orcidlink}
\usepackage{array}
\usepackage{epigraph}
\usepackage{multirow}
\usepackage{pifont}
\usepackage{float}
\usepackage{url}            

\usepackage{amssymb}
\usepackage{pifont}

\usepackage{CJKutf8}
\usepackage{longtable}

\definecolor{colorcommentbg_pkprompt}{HTML}{5D8AA8}
\definecolor{colorcommentframe_pkprompt}{HTML}{0093AF}
\newenvironment{pkprompt}[1][]{
	\begin{tcolorbox}[adjusted title={Pair-wise Comparison Prompt for GPT4V(ision)}, fonttitle={\bfseries\footnotesize}, fontupper=\scriptsize, colback={colorcommentbg_pkprompt!30}, colframe={colorcommentframe_pkprompt!80},coltitle={white},#1]
}{\end{tcolorbox}}

\newcommand{\dalle}{DALL$\cdot$E3\xspace}

\newcommand{\ourname}{\textsc{FontStudio}\xspace}
\newcommand{\ourbenchmark}{\textsc{GenerativeFont}\xspace}

\newlength\savewidth\newcommand\shline{\noalign{\global\savewidth\arrayrulewidth
  \global\arrayrulewidth 1pt}\hline\noalign{\global\arrayrulewidth\savewidth}}
\newcommand{\tablestyle}[2]{\setlength{\tabcolsep}{#1}\renewcommand{\arraystretch}{#2}\centering\footnotesize}

\newcommand{\model}{Shape-Adaptive Diffusion Model\xspace}

\newcommand{\maskclip}{M-CLIP-Int\xspace}
\newcommand{\maskclipout}{M-CLIP-Ext\xspace}
\newcommand{\sgm}{Shape-adaptive Generation Model\xspace}
\newcommand{\srm}{Shape-adaptive Refinement Model\xspace}

\makeatletter
\def\@fnsymbol#1{\ensuremath{\ifcase#1\or \dagger\or \ddagger\or
\mathsection\or \mathparagraph\or \|\or **\or \dagger\dagger
\or \ddagger\ddagger \else\@ctrerr\fi}}
\makeatother

\begin{document}

\title{FontStudio: Shape-Adaptive Diffusion Model for Coherent and Consistent Font Effect Generation}

\titlerunning{FontStudio: Shape-Adaptive Diffusion Model}

\author{{\normalsize Xinzhi Mu \quad Li Chen \quad Bohan Chen\thanks{Intern at Microsoft.}
 \quad Shuyang Gu \\ Jianmin Bao \quad\; Dong Chen \quad\; Ji Li\quad\; Yuhui Yuan}}

\authorrunning{Xinzhi Mu et al.}

\institute{Microsoft \\
\email{\{xinzhimu,yuyua\}@microsoft.com}\\
\url{https://font-studio.github.io/}}

\onecolumn{%
\renewcommand\onecolumn[1][]{#1}%
\maketitle
\begin{center}
\begin{minipage}[t]{1\linewidth}
\includegraphics[height=51mm]{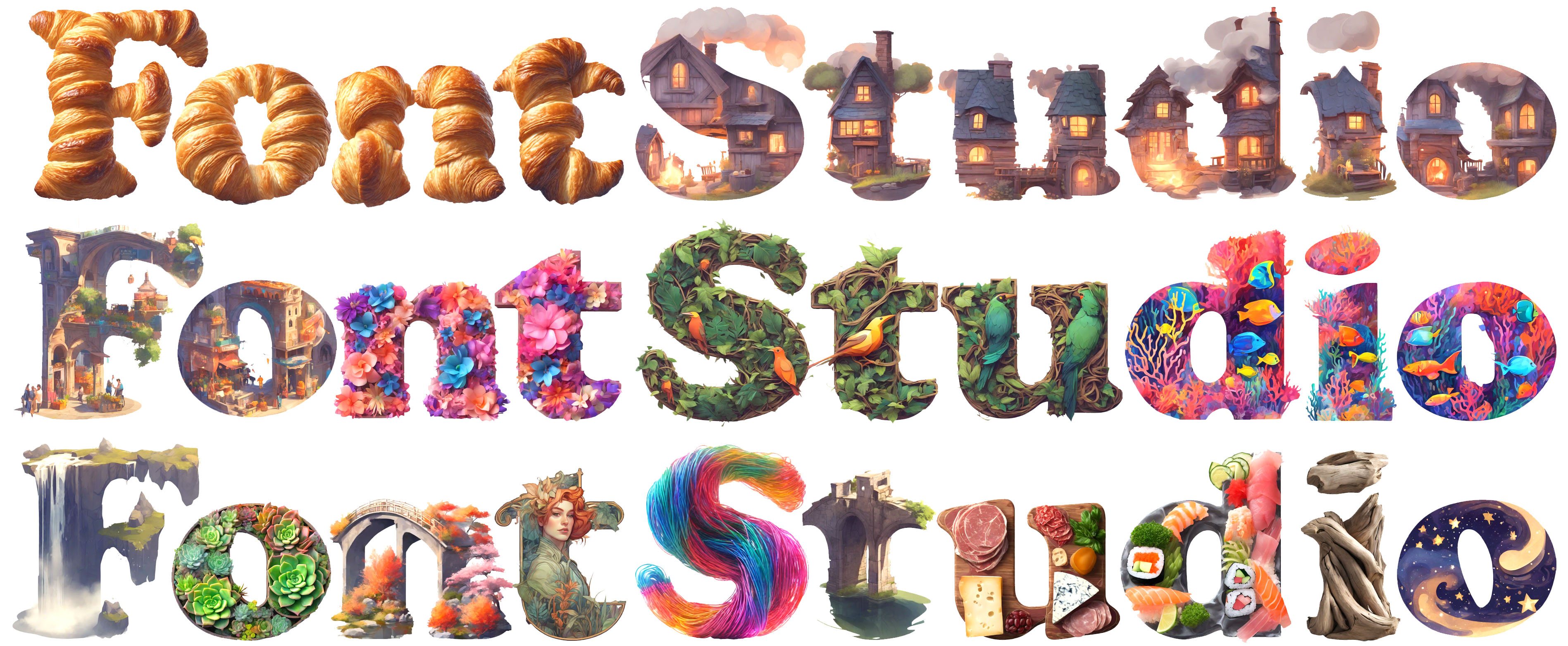}
\vspace{-6mm}
\captionof{figure}{\small{Illustrating the font effect generation results by our \ourname system. We observe that most concepts are generated in adherence to complex font shapes adaptively. We also notice a coherent 3D structure and depth effect. Refer to the supplementary for a detailed prompt of these generative font effects.}}
\vspace{-4mm}
\label{fig:teaser}
\end{minipage}
\end{center}
}

\begin{abstract}
Recently, the application of modern diffusion-based text-to-image generation models for creating artistic fonts, traditionally the domain of professional designers, has garnered significant interest. Diverging from the majority of existing studies that concentrate on generating artistic typography, our research aims to tackle a novel and more demanding challenge: the generation of text effects for multilingual fonts. This task essentially requires generating coherent and consistent visual content within the confines of a font-shaped canvas, as opposed to a traditional rectangular canvas. To address this task, we introduce a novel shape-adaptive diffusion model capable of interpreting the given shape and strategically planning pixel distributions within the irregular canvas. To achieve this, we curate a high-quality shape-adaptive image-text dataset and incorporate the segmentation mask as a visual condition to steer the image generation process within the irregular-canvas. This approach enables the traditionally rectangle canvas-based diffusion model to produce the desired concepts in accordance with the provided geometric shapes. Second, to maintain consistency across multiple letters, we also present a training-free, shape-adaptive effect transfer method for transferring textures from a generated reference letter to others. The key insights are building a font effect noise prior and propagating the font effect information in a concatenated latent space.
The efficacy of our \ourname system is confirmed through user preference studies, which show a marked preference (78\% win-rates on aesthetics) for our system even when compared to the latest unrivaled commercial product, Adobe Firefly\footnote{\url{https://firefly.adobe.com/generate/font-styles}}.
\end{abstract}

\section{Introduction}
\label{sec:intro}

Recently, models based on diffusion techniques for text-to-image generation have achieved significant success in rendering photorealistic images on standard rectangular canvases~\cite{podell2023sdxl,DeepFloyd_IF,dalle3paper}. Many follow-up efforts have built many generation-driven exciting applications like subject-driven image generation and spatial conditional image generation. For instance, ControlNet~\cite{zhang2023adding} offers a powerful method for integrating spatial conditioning controls, such as edges, depth, segmentation, and more, into pre-trained text-to-image diffusion models, enhancing their versatility and application range.

\begin{figure}[htbp] 
\vspace{-2mm}
\tiny
\centering
\begin{minipage}{.68\textwidth}
\resizebox{0.98\linewidth}{!}
{
\begin{tabular}{ccccccc}
    Mask & {CN-Depth} & CN-Canny & SDXL & SDXL-Inpaint & Adobe-Firefly& \ourname \\
\includegraphics[width=0.155\textwidth]{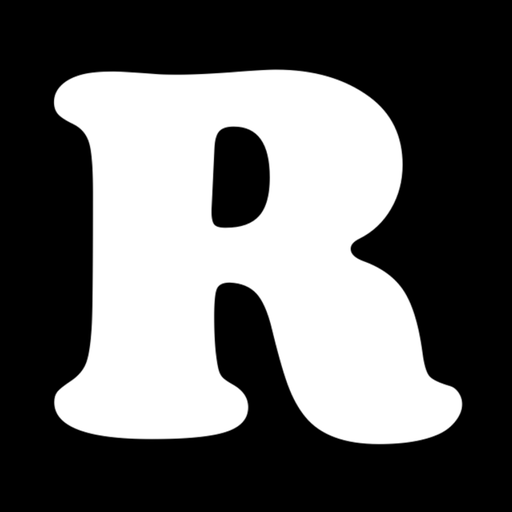}&
\includegraphics[width=0.155\textwidth]{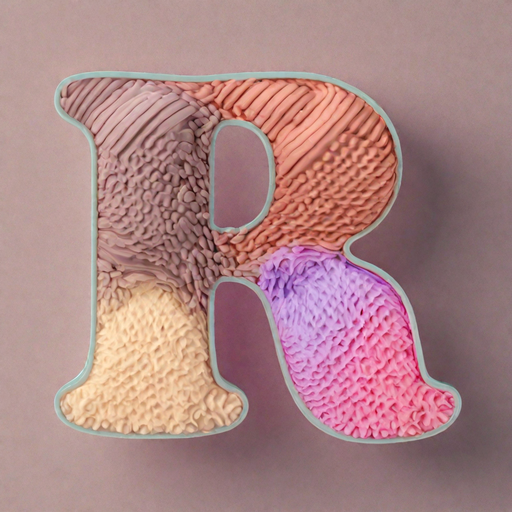}&
\includegraphics[width=0.155\textwidth]{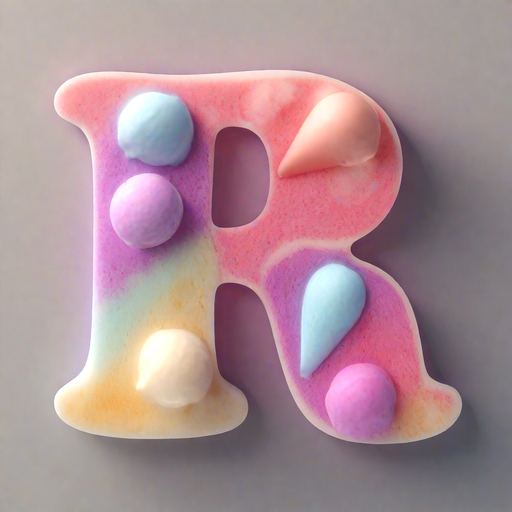}&
\includegraphics[width=0.155\textwidth]{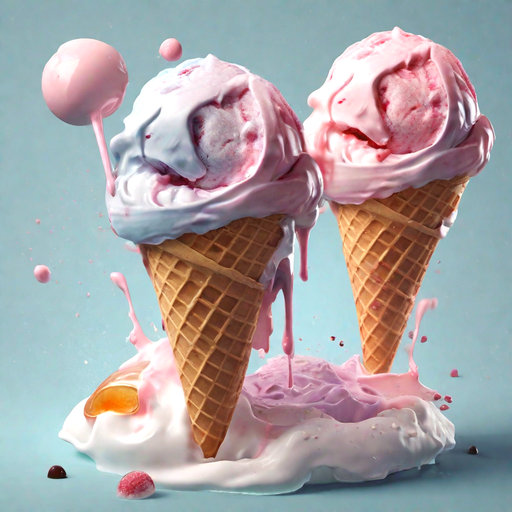}&
\includegraphics[width=0.155\textwidth]{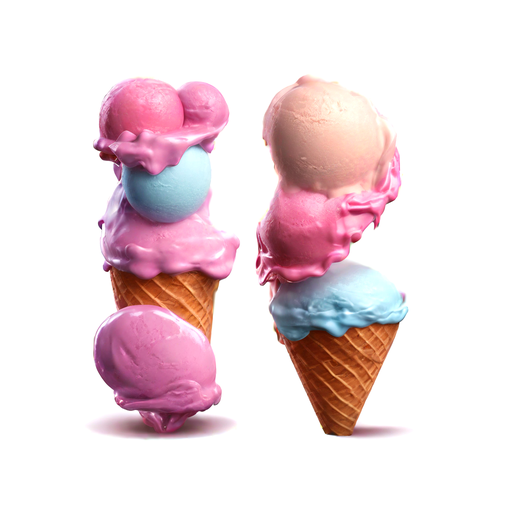}&
\includegraphics[width=0.155\textwidth]{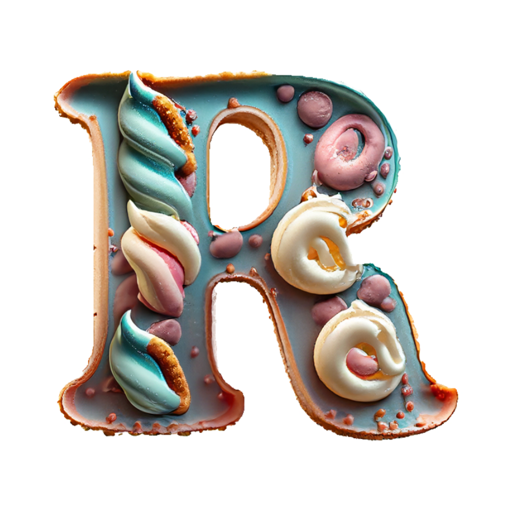}&
\includegraphics[width=0.155\textwidth]{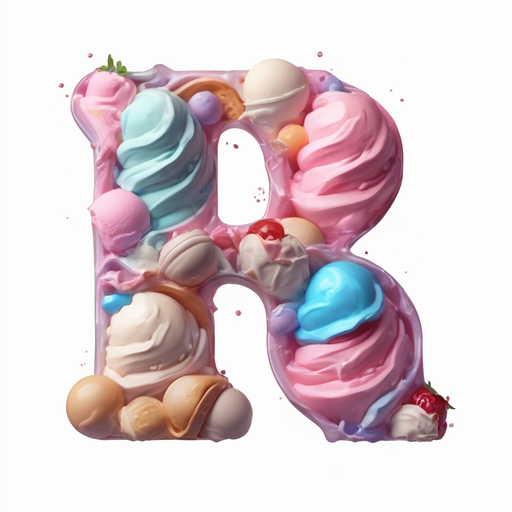}\\
\includegraphics[width=0.155\textwidth]{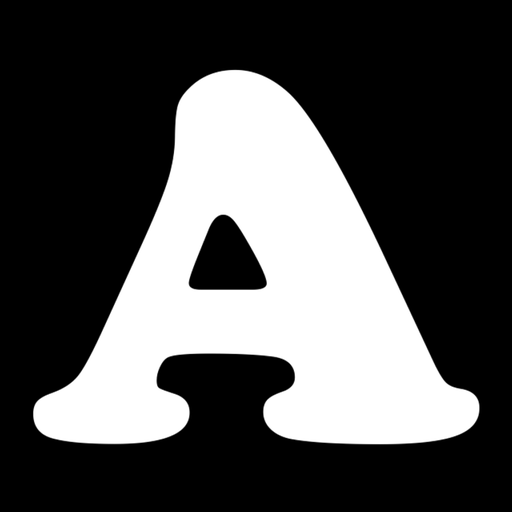}&
\includegraphics[width=0.155\textwidth]{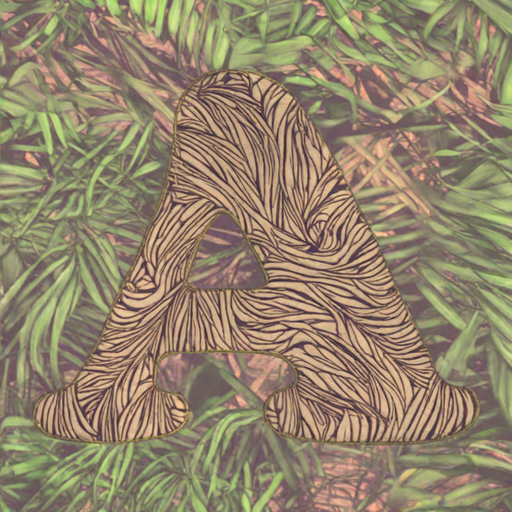}&
\includegraphics[width=0.155\textwidth]{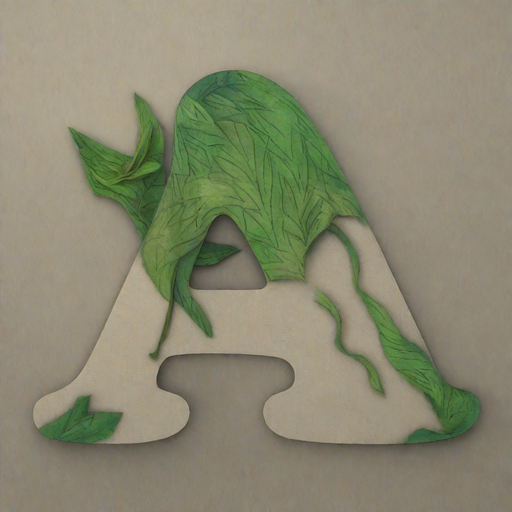}&
\includegraphics[width=0.155\textwidth]{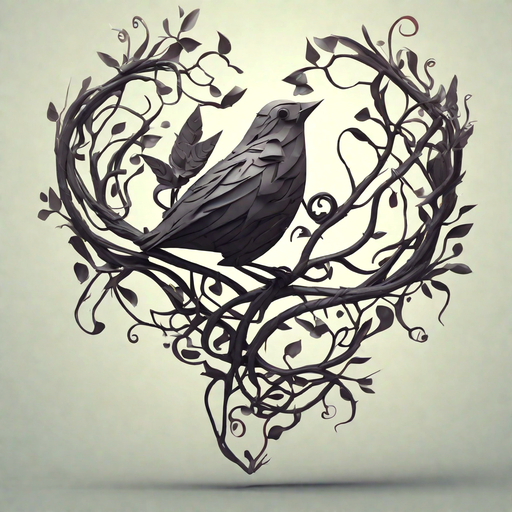}&
\includegraphics[width=0.155\textwidth]{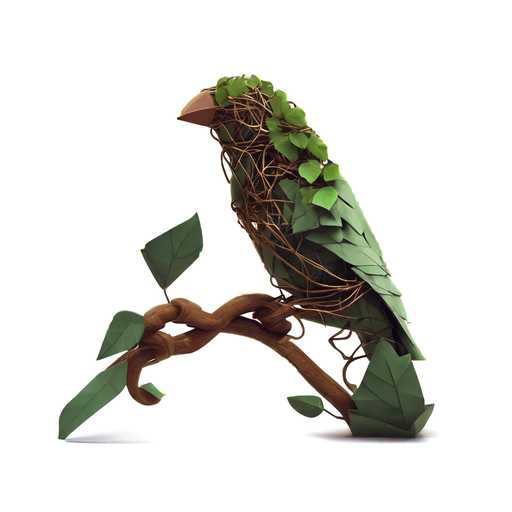}&
\includegraphics[width=0.155\textwidth]{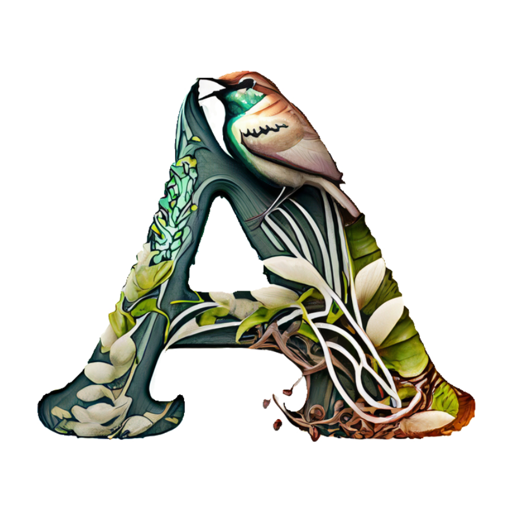}&
\includegraphics[width=0.155\textwidth]{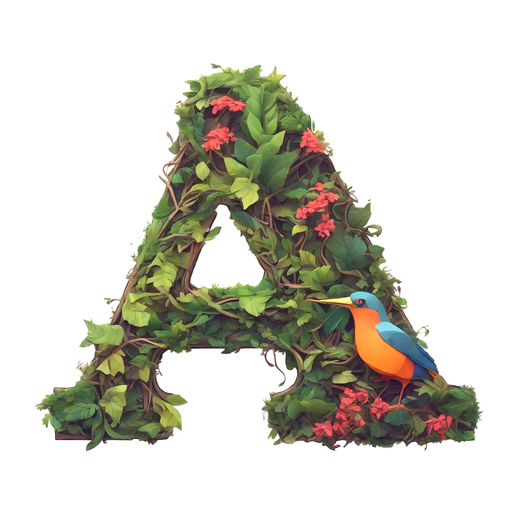}
\end{tabular}
}
\caption{\small \textbf{Comparison with conventional diffusion models designed for rectangular canvas.} Most of these methods struggle to generate the appealing visual content within font-shaped canvas. For ControlNet (CN), we find treating the font mask as depth or computing the canny edge map based on font mask suffers various artifacts. Our \ourname generates much better results in general.}
\label{fig:rectangle_canva_results}
\end{minipage}
\hfill
\begin{minipage}{.29\textwidth}
\centering
\includegraphics[width=0.99\textwidth]{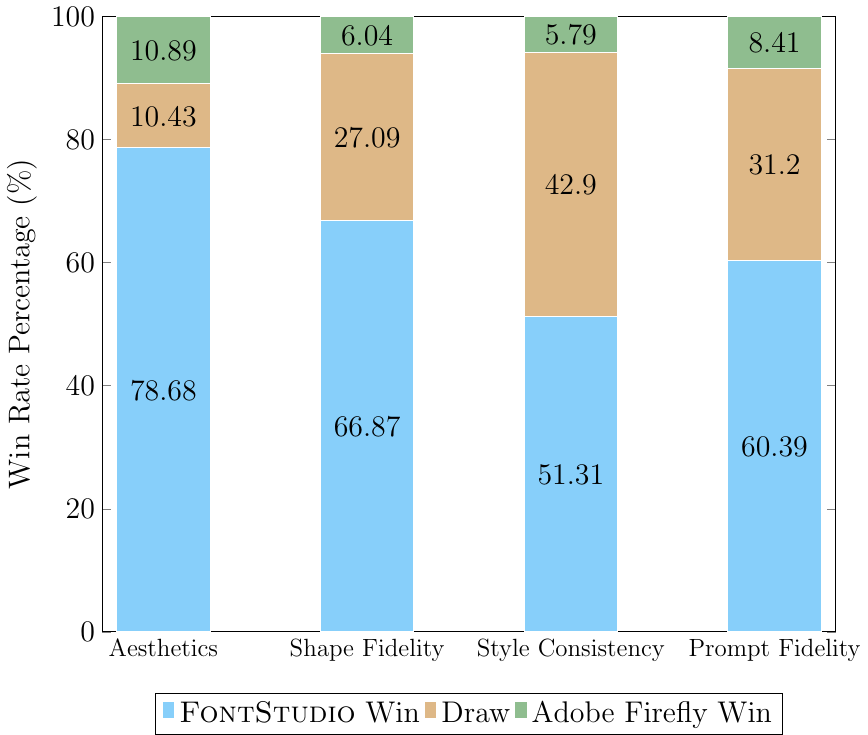}
\caption{\footnotesize{\textbf{FontStudio vs. Adobe Firefly.} Win-rates accessed by human evaluator preferences in font effect generation.}}
\label{fig:user_study}
\end{minipage}
\vspace{-3mm}
\end{figure}

Despite these advancements, the focus predominantly remains on rectangular canvases, leaving the potential for image generation on non-standard, arbitrarily shaped canvases largely untapped.
The task of creative font effect generation essentially requires generating visual contents in non-regular and complex-shaped canvas.
It demands not only synthesizing semantic objects or concepts aligning with arbitrary user prompts but also a deep understanding of the geometric shapes of the font canvas.
In essence, the visual elements produced must be precisely positioned within the irregular-canvas to ensure visual harmony while also ensuring faithful generation within the specific font canvas following the given text prompt. Our empirical analysis, illustrated in Figure~\ref{fig:rectangle_canva_results}, demonstrates the outcomes of directly utilizing conventional diffusion models, including ControlNet, SDXL, and SDXL-Inpainting model, designed for rectangular canvases. From this analysis, it becomes evident that simply adapting models intended for rectangular canvases to generate visual content for the diverse array of font shapes presents a significant and largely uncharted challenge in the field.

To bridge the gap between traditional rectangle-canvas-based diffusion models and the intricate task of comprehending font shapes for font effect generation, we propose an innovative and potent shape-adaptive diffusion model. This model excels in producing high-quality visual content that conforms to any given shape, encompassing multilingual font outlines and even more intricate patterns such as fractal-structured snowflakes.
The key idea is to build a high-quality shape-adaptive triplet training data and each instance consists of \{irregular-canvas, irregular-image, text prompt\}  and then train a conditional diffusion model to generate the visual contents within the irregular-canvas.
To maintain compatibility with pre-trained diffusion models and ensure efficient training, we choose a rectangular canvas to serve as a placeholder, accommodating both the irregularly shaped canvas and the corresponding irregular image.

The task of generating font effects requires preserving effect consistency across multiple irregular canvases. Merely using the diffusion model in isolation often results in inconsistent outcomes. To address this challenge, we introduce a novel, training-free effect transfer method that combines the effect of a reference letter with the shape mask of a target letter. This method leverages a font effect noise prior to ensure font effect consistency and propagates the reference style and texture from the source to the target image in a concatenated latent space. Our empirical results demonstrate that this approach can effectively serve as a powerful tool for transferring effects or styles.

Last, we established the \ourbenchmark benchmark to facilitate a comprehensive evaluation of our methodologies across various dimensions. The results from a user study, depicted in Figure~\ref{fig:user_study}, when benchmarked against Adobe Firefly—the leading font effect generation system—reveal a surprising outcome. Our \ourname system markedly outperforms Adobe Firefly in several key areas. Specifically, thanks to our shape-adaptive generation approach, we observed a remarkable improvement in both shape fidelity and overall aesthetics, with our system achieving win rates of 78.68\% vs. 10.89\% in aesthetics and 66.87\% vs. 6.04\% in shape fidelity.
While \ourname secures these promising achievements, we continue to thoroughly investigate the system's limitations and engage in discussions on emerging challenges that beckon attention from the broader research community.

\section{Related Work}

\noindent\textbf{Artistic Font Generation.}
Previous research has explored various facets of font-related tasks, with studies such as \cite{campbell2014learning, balashova2019learning, wang2021deepvecfont} concentrating on font creation. Other methods, including GAN-based approaches \cite{azadi2018multi, jiang2019scfont, gao2019artistic, Yang_2019_ICCV}, stroke-based techniques \cite{berio2022strokestyles}, and statistical approaches \cite{yang2017awesome, yang2018context, yang2018context2}, aim to transfer existing image styles onto font images. Additionally, research on semantic font typography \cite{kwan2016pyramid, saputra2019improved, chen2019manufacturable, zhang2022creating, xu2007calligraphic, zou2016legible} investigates 2D collage generation and reverse challenges, while \cite{tendulkar2019trick, iluz2023word, tanveer2023ds} focus on modifying characters for thematic expression without sacrificing readability. There are also frameworks for glyph design, either leveraging existing assets \cite{zhang2017synthesizing} or large language models \cite{he2023wordart}. Anything to Glyph \cite{wang2023anything} parallels our study by suggesting alterations to location features in the diffusion model's denoising phase. However, this method often produces noticeable shadows in both the foreground and background, restricting its utility in further design applications and failing to ensure character consistency throughout generation.
Unlike these existing studies, we focus on generating text effects for multilingual fonts, aiming to produce coherent and consistent visual content within the confines of a font-shaped canvas.

\noindent\textbf{Diffusion-based Image Synthesis and Attention.}
The landscape of text-to-image generation has seen considerable growth in recent times \cite{nichol2021glide, rombach2022high, chang2023muse}, with diffusion-based methodologies \cite{saharia2205photorealistic, rombach2022high, podell2023sdxl} at the forefront, pushing the boundaries of image synthesis quality. The scope of investigation has broadened from straightforward text-to-image conversions to encompass a variety of intricate image applications, including conditional image generation \cite{zhang2023adding, voynov2023sketch}, image inpainting \cite{saharia2022palette, avrahami2023blended}, image-to-image translation \cite{nichol2021glide, rombach2022high, chang2023muse}, image editing \cite{couairon2022diffedit, huberman2023edit}, and tailored generation \cite{hu2021lora, ruiz2023dreambooth, ruiz2023hyperdreambooth, tewel2023key}. It is noteworthy that these explorations predominantly take place on standard rectangular canvases.
The integration of attention mechanisms in diffusion models has spurred a variety of research in areas like image editing \cite{chefer2023attend, patashniklocalizing, cao2023masactrl, mokady2023null, parmar2023zero, epstein2024diffusion}. Recent works such as \cite{alaluf2023cross, park2024shape} explore attention for style transfer, with StyleAligned \cite{park2024shape} closely aligning with our shape-adaptive effect transfer's goals of achieving stylistic consistency through attention-directed generation using reference images. We empirically show that our method performs better in delivering stylistically coherent generated images while preserving diversity.

\section{Approach}
First, we illustrate the definition and mathematical formulation of the font effect generation task, delve into the primary challenges associated with this task, and outline the foundational insights guiding our methodology. Second, we introduce the key contribution of this work, namely, a shape-adaptive diffusion model, designed to produce visual content on canvases of any shape.
Last, we detail the implementation of our shape-adaptive effect transfer method, which utilizes font effect noise prior and font effect propagation to achieve our objectives.

\subsection{Preliminary}
\label{subsec:premininary}

We use the subscript $\;\widehat{}\;$ to indicate that the given tensor has a non-rectangular and irregular spatial shape. For example, $\mathbf{X}$ represents a tensor with a rectangular spatial shape like $h \times w$, while $\widehat{\mathbf{X}}$ denotes a tensor of irregular shape with variable dimensions.

\noindent\textbf{Definition of font effect generation.}
Given a target font effect text prompt \(\mathbf{T}\) and a sequence of irregular font shape canvases \{\(\widehat{\mathbf{M}}_i | i=1,...,n\)\} corresponding to a sequence of letters, the objective is to build a set-to-set mapping function \(f(\cdot)\) that can generate a set of coherent and consistent font effect images \{\(\widehat{\mathbf{I}}_i | i=1,...,n\)\} of the same shape as the given irregular font-shape canvases \{\(\widehat{\mathbf{M}}_i | i=1,...,n\)\} accordingly. We illustrate the mathematical formulation of font effect generation process as follows:
\begin{equation}
\{\widehat{\mathbf{I}}_i \;|\; i=1,...,n\} = f(\{\widehat{\mathbf{M}}_i \;|\; i=1,...,n\}\;|\;\mathbf{T}),
\end{equation}
where we can also use different font effect text for each mask separately. We propose to access the font effect generation quality from the following four critical aspects:
\begin{itemize}
\item \emph{Aesthetics}: Each generated image \(\widehat{\mathbf{I}}_i\) should be visually attractive.
\item \emph{Font Shape Fidelity}: While an exact match isn't necessary, each \(\widehat{\mathbf{I}}_i\) should closely resemble its original font shape \(\widehat{\mathbf{M}}_i\).
\item \emph{Font Style Consistency}: \(\widehat{\mathbf{I}}_i\) should exhibit a coherent style for any other image \(\widehat{\mathbf{I}}_j\), presenting as a unified design.
\item \emph{Prompt Fidelity}: Every \(\widehat{\mathbf{I}}_i\) must adhere to the provided target effect prompt.
\end{itemize}

\vspace{1mm}
\noindent\textbf{Primary challenges.}
The first key challenge in font effect generation is ensuring that the generated visual objects are positioned creatively and coherently on the font-shaped canvas. We have already shown that the results from simply applying diffusion models designed for rectangular canvases are far from satisfactory, as demonstrated in Figure~\ref{fig:rectangle_canva_results}. The second challenge involves maintaining font shape fidelity, as the primary goal of generative fonts is to convey messages creatively. Additionally, ensuring consistent font effects across different letters is also a non-trivial and challenging task, considering the canvas shapes vary significantly among different letters.

\vspace{1mm}
\noindent\textbf{Formulation of our framework.}
To address the above challenges, we first reformulate the font effect generation task into the combination of two sub-tasks including \emph{font effect generation for a reference letter} and \emph{font effect transfer from reference letter to each other letter}.
The mathematical formulation is summarized as follows:
\begin{align}
\label{eq:irregular_shape} 
 \widehat{\mathbf{I}}_{\rm{ref}} &= {g}(\widehat{\mathbf{M}}_{\rm{ref}}\;|\;\mathbf{T}),\\
\label{eq:effect_transfer}   
\widehat{\mathbf{I}}_i &= {h}(\widehat{\mathbf{M}}_i\;|\;\mathbf{T},\; \widehat{\mathbf{M}}_{\rm{ref}}, \;\widehat{\mathbf{I}}_{\rm{ref}}),\; \rm{i}\in\{1, \cdots, n\},
\end{align}
where we use the function \({g}(\cdot)\) to perform font effect generation based on a single irregular reference canvas, denoted as $\widehat{\mathbf{M}}_{\rm{ref}}$. The function \({h}(\cdot)\) is used to generate consistent font effects, conditioned on the previously generated reference font effect image $\widehat{\mathbf{I}}_{\rm{ref}}$, the reference font mask $\widehat{\mathbf{M}}_{\rm{ref}}$, and the current font mask $\widehat{\mathbf{M}}_{i}$. We choose the same reference letter mask for all font effect transfer letters.
To implement these two critical functions, we proposed a \model marked as \({g}(\cdot)\) and a Shape-adaptive Effect Transfer together with \model marked as \({h}(\cdot)\).
We will explain the details in the following discussion.

\vspace{-2mm}
\subsection{\model}
\label{sec:model}
The key challenge in font effect generation arises from the gap between most existing diffusion models, which are trained on rectangular canvases, and the requirement of this task for visual content creation capability on any given irregularly shaped canvas. To close this critical gap, we introduce a shape-adaptive diffusion model that is capable of performing visual content creation on any irregularly shaped canvas as function \({g}(\cdot)\).

We follow the mathematical formulations outlined in Equation~\ref{eq:irregular_shape} and utilize the transformation function \({g}(\cdot)\), which is applied to the irregular canvas \(\widehat{\mathbf{M}}_i\) conditioned on a given user prompt \(\mathbf{T}\), to represent the shape-adaptive diffusion model. The output of the function \({g}(\cdot)\) is essentially an image \(\widehat{\mathbf{I}}_i\) with an irregular shape. Given that directly processing irregular canvases of varying resolutions presents several non-trivial challenges in training standard diffusion models, we propose rasterizing and positioning the irregular canvas mask within a rectangular placeholder, as \(\mathbf{M} = \mathsf{Rasterize}(\widehat{\mathbf{M}})\). Essentially, \(\mathbf{M}\) is the binary rasterized form of \(\widehat{\mathbf{M}}\) where the pixels inside \(\widehat{\mathbf{M}}\) are with 1 and the other pixels are with 0. Additionally, we utilize a rectangular image \(\mathbf{I}\) to encapsulate the irregular font effect image \(\widehat{\mathbf{I}}\) and include an irregular alpha mask layer \(\mathbf{M}_\mathbf{I}\) to eliminate the regions outside the irregular canvas.
Given the irregular shaped canvas mask and image encapsulated within rectangle ones,
we reformulate the original  Equation~\ref{eq:irregular_shape} as follows:
\begin{align}
\label{eq:reformed_g}
{\mathbf{I}}, \mathbf{M}_{\mathbf{I}} &= \bar{g}({\mathbf{M}}\;|\;\mathbf{T})
\end{align}
where the predicted alpha mask layer \(\mathbf{M}_{\mathbf{I}}\) is different from the input conditional font mask ${\mathbf{M}}$ and it is necessary to ensure coherent and creative effects along the boundary regions.
With the alpha mask prediction, we also avoid the necessity to use additional segmentation model to handle the artifacts outside the font-shaped canvas. We elucidate the key that differentiating the refined alpha mask from the conditional canvas mask is achieved through canvas mask augmentation during the training of the subsequent shape-adaptive diffusion model.

\begin{figure}[h]
\centering
\includegraphics[width=1.0\textwidth]{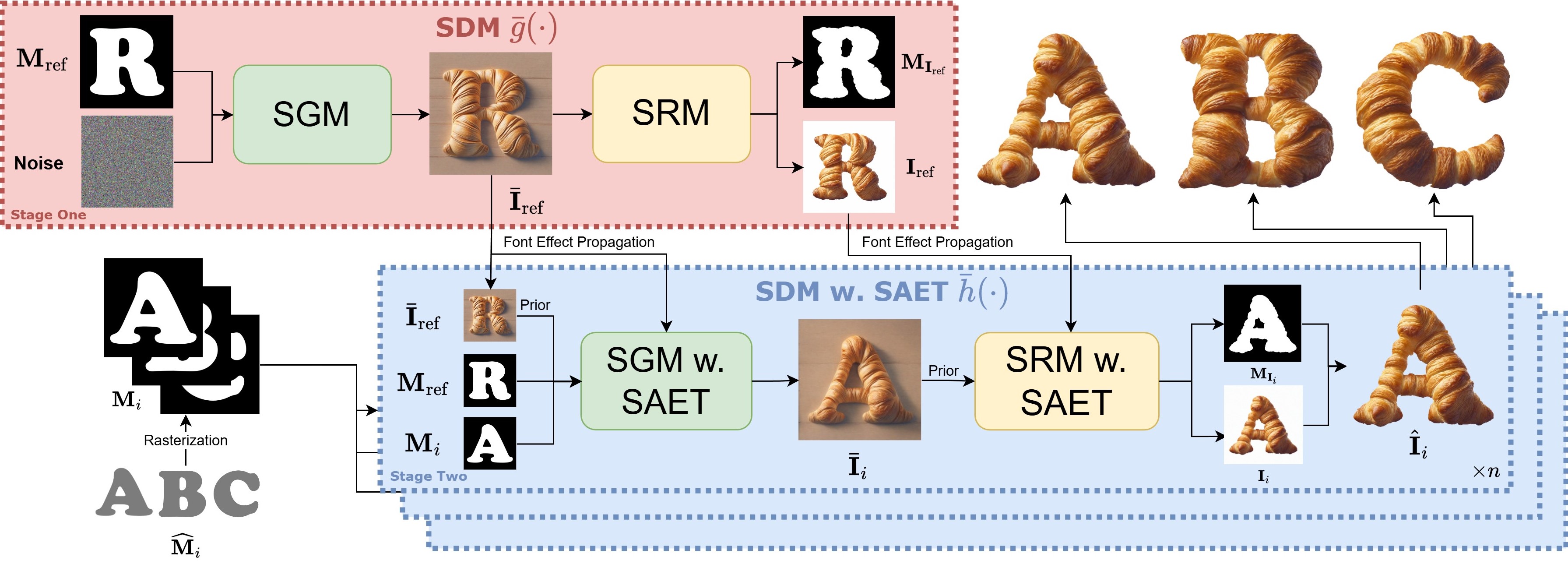}
\includegraphics[width=1.0\textwidth]{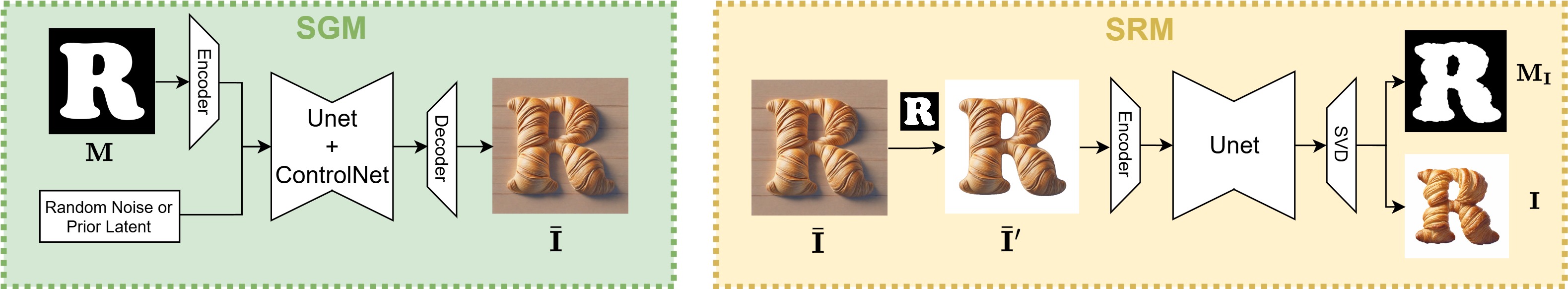}
\caption{\small \textbf{Overall framework of our approach.} 
The shape-adaptive diffusion model (SDM) consists of two components: the shape-adaptive generation model (SGM) and the shape-adaptive refinement model (SRM). The SGM generates content within a rasterized shape, whereas the SRM refines content edges and produces a refined shape alpha mask using our shape-adaptive VAE decoder (SVD). In stage one, we use SDM to generate reference images and in stage two, by employing shape-adaptive effect transfer (SAET), we transfer the style of reference images to target images to ensure style consistency between $\widehat{\mathbf{I}}_i$. Prior indicates font effect noise prior used in SAET.}
\label{fig:overall_framework}
\end{figure}

Our shape-adaptive diffusion model consists of two sub-models: a shape-adaptive generation model followed by a shape-adaptive refinement model. The shape-adaptive generation model, dubbed SGM, primarily generates content relevant to the prompt within a designated region, utilizing \(\mathbf{M}\). Following this, the shape-adaptive refinement model (SRM) takes over, aiming to enhance the initial results by creating an image \(\mathbf{I}\) with more defined and natural edges, along with the corresponding mask \(\mathbf{M}_\mathbf{I}\) for the generated image \(\mathbf{I}\).
Figure~\ref{fig:overall_framework} illustrates the entire framework of of our approach.

\vspace{1mm}
\noindent\textbf{\sgm.}
Training a shape-adaptive generation model is non-trivial, and we face two key challenges. The first is the lack of high-quality training data that aligns text with images encapsulated within an irregularly shaped canvas. The second challenge arises from the default self-attention and cross-attention schemes, which directly map text information across the entire rectangular canvas. This approach inadvertently allows for visual content generation in regions outside the irregularly shaped canvas, which is also rasterized into a rectangle, thereby diminishing the effectiveness of targeted content generation within the desired canvas region. To overcome these challenges, we propose two key contributions: constructing high-quality shape-adaptive image-text data and implementing a shape-adaptive attention scheme. We elaborate more details on these two techniques in the following discussion.

\begin{figure}[htbp] 
\vspace{-2mm}
\tiny
\centering
\resizebox{0.99\linewidth}{!}
{
\begin{tabular}{ccccccccc}
\tiny
\tiny \raisebox{0.01\textwidth}{\rotatebox[origin=l]{90}{\dalle}} &
\includegraphics[width=0.12\textwidth]{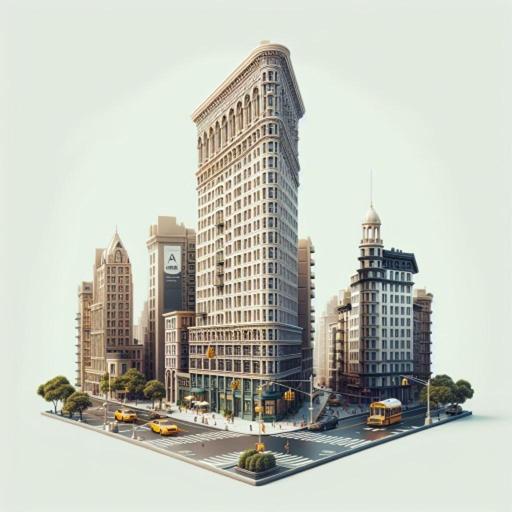}&
\includegraphics[width=0.12\textwidth]{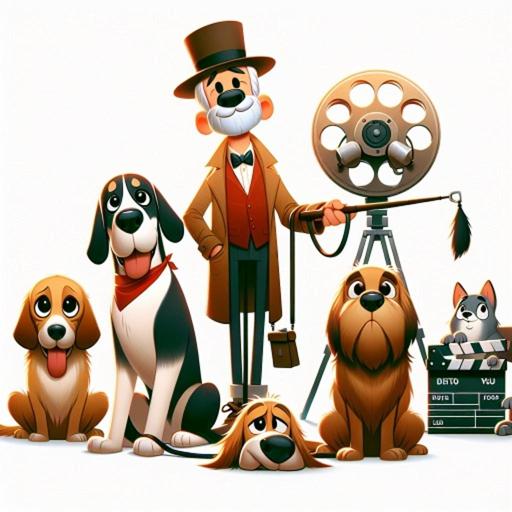}&
\includegraphics[width=0.12\textwidth]{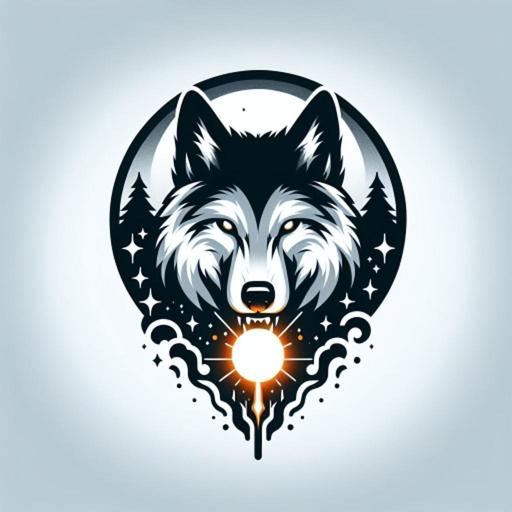}&
\includegraphics[width=0.12\textwidth]
{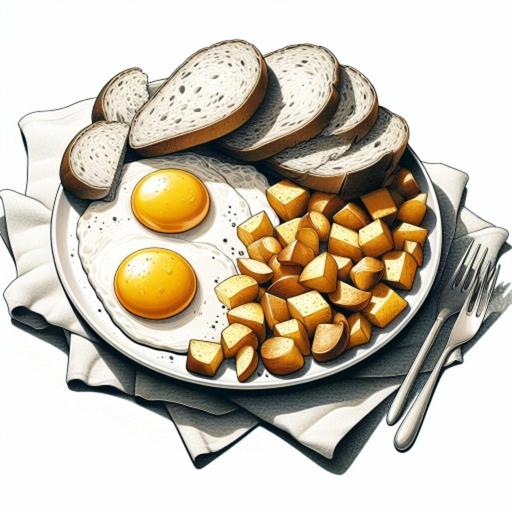}&
\includegraphics[width=0.12\textwidth]{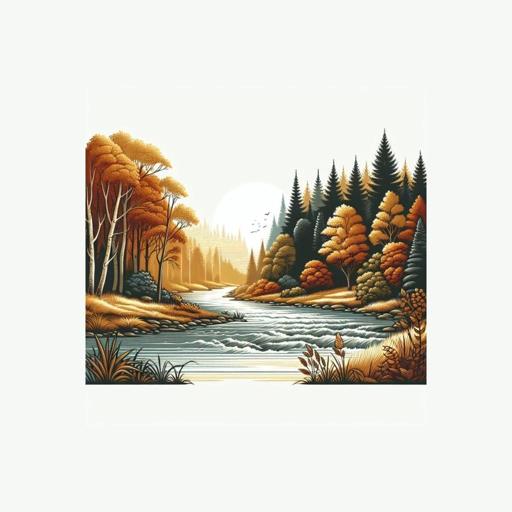}&
\includegraphics[width=0.12\textwidth]{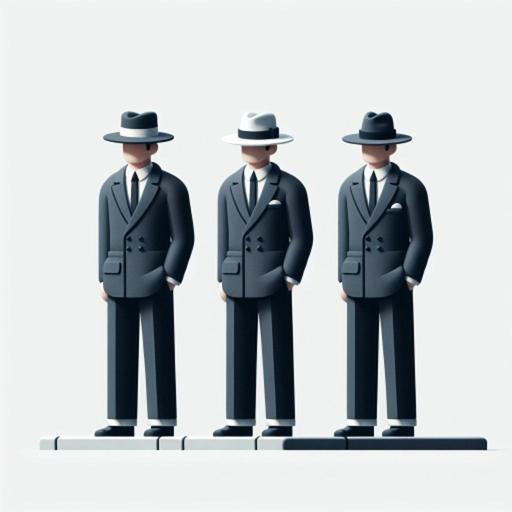}&
\includegraphics[width=0.12\textwidth]{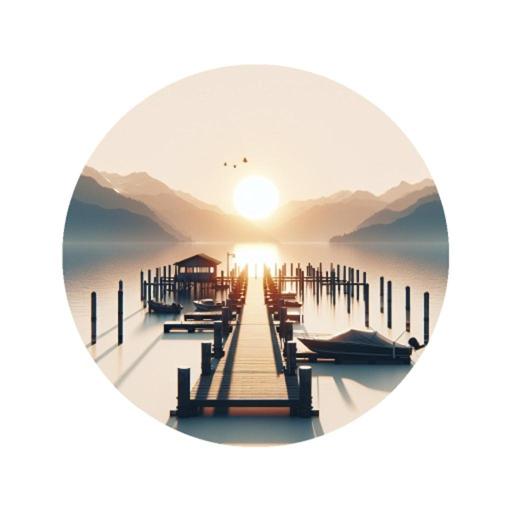}&
\includegraphics[width=0.12\textwidth]{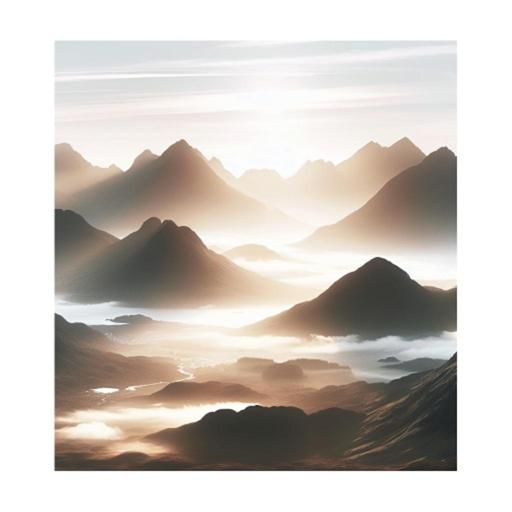}\\
\tiny \raisebox{0.025\textwidth}{\rotatebox[origin=l]{90}{Mask}} &
\includegraphics[width=0.12\textwidth]{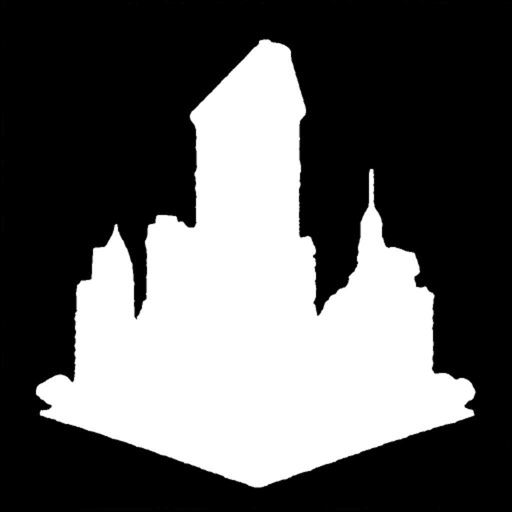}&
\includegraphics[width=0.12\textwidth]{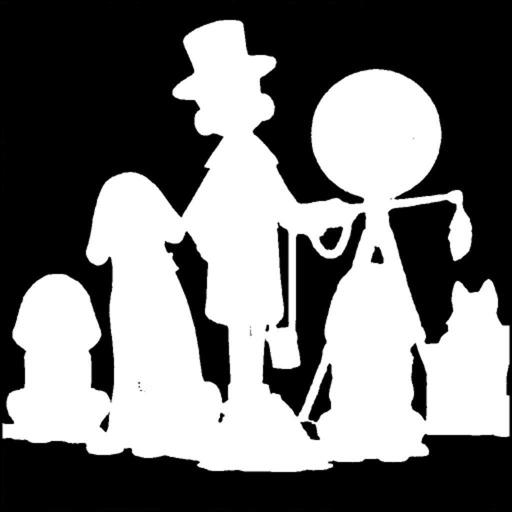}&
\includegraphics[width=0.12\textwidth]{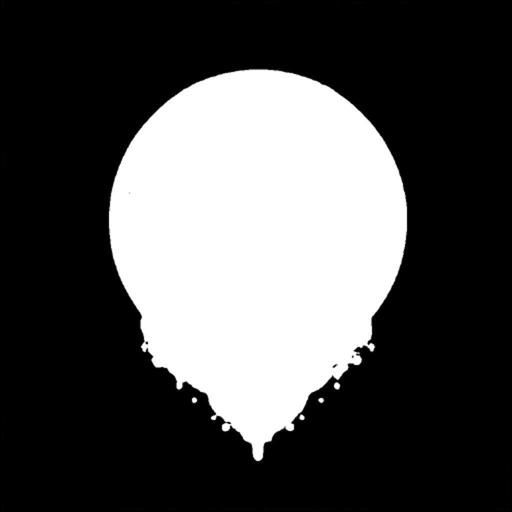}&
\includegraphics[width=0.12\textwidth]
{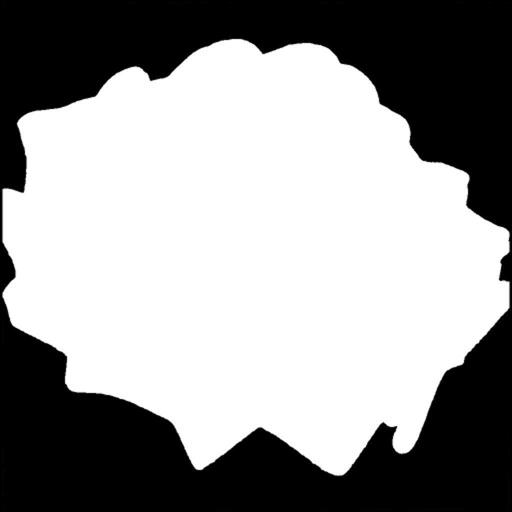}&
\includegraphics[width=0.12\textwidth]{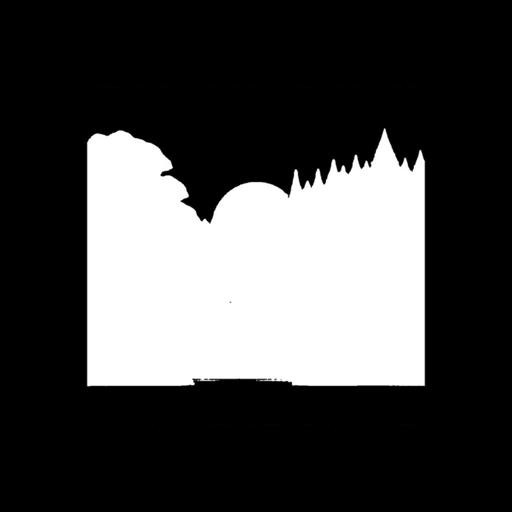}&
\includegraphics[width=0.12\textwidth]{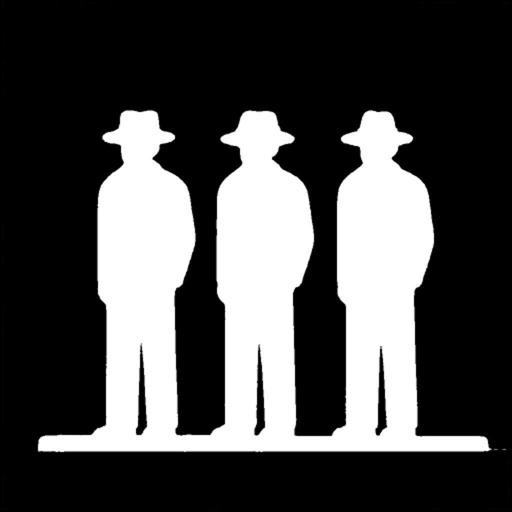}&
\includegraphics[width=0.12\textwidth]{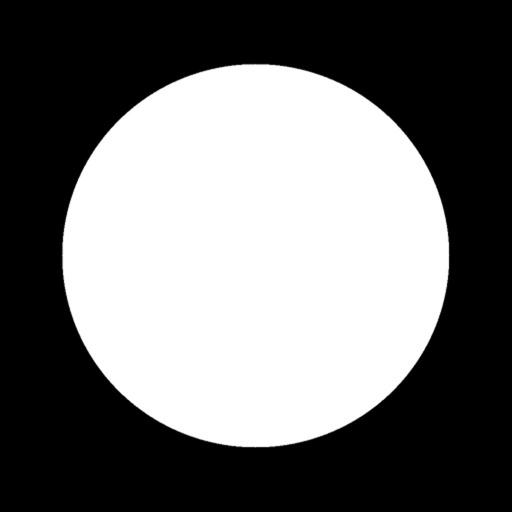}&
\includegraphics[width=0.12\textwidth]{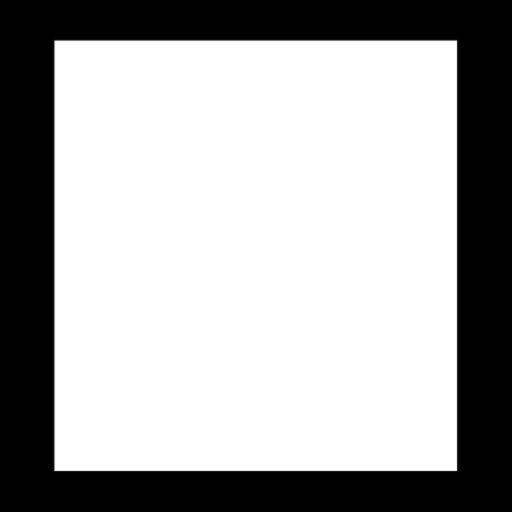}\\
\tiny \raisebox{0.01\textwidth}{\rotatebox[origin=l]{90}{Aug-Mask}} &
\includegraphics[width=0.12\textwidth]{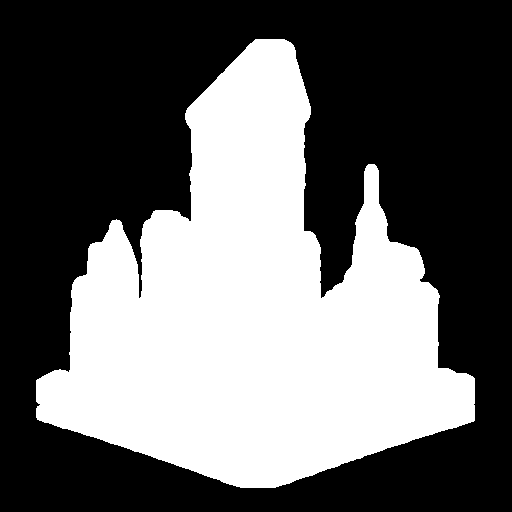}&
\includegraphics[width=0.12\textwidth]{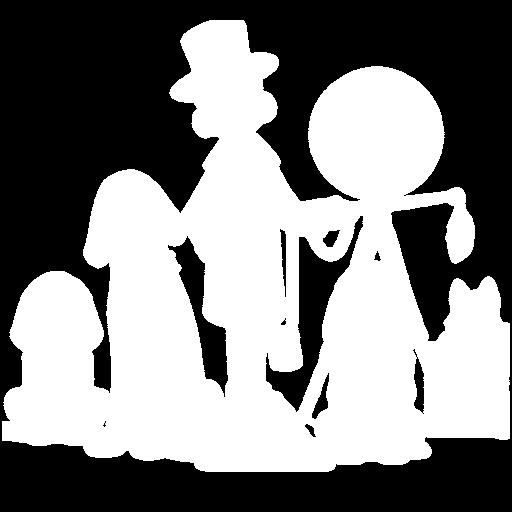}&
\includegraphics[width=0.12\textwidth]{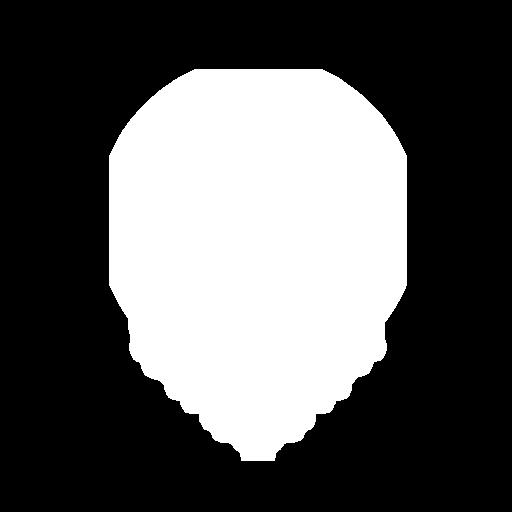}&
\includegraphics[width=0.12\textwidth]
{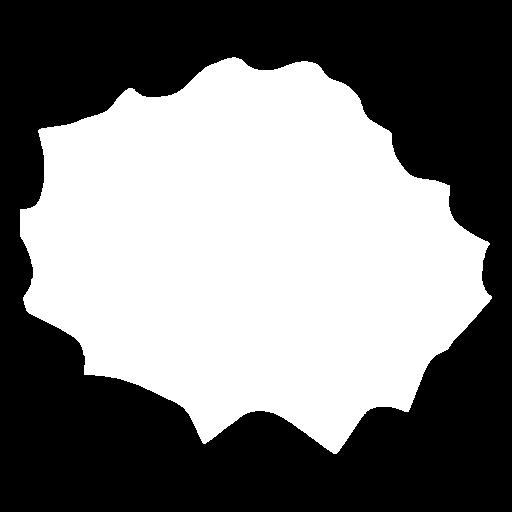}&
\includegraphics[width=0.12\textwidth]{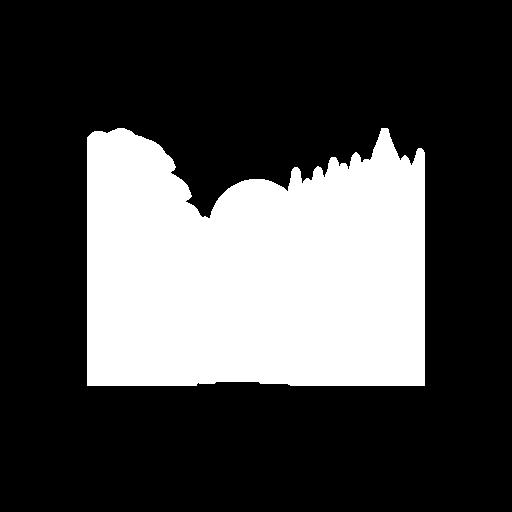}&
\includegraphics[width=0.12\textwidth]{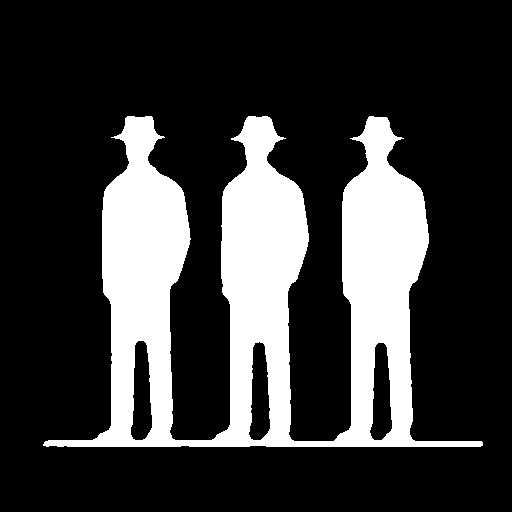}&
\includegraphics[width=0.12\textwidth]{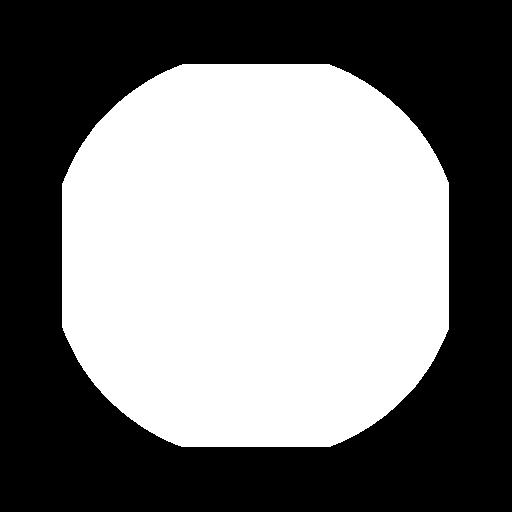}&
\includegraphics[width=0.12\textwidth]{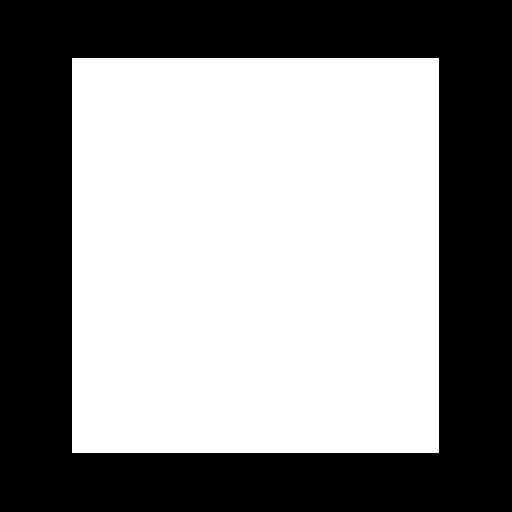}\\
\end{tabular}
}
\caption{\small Illustrating examples of our shape-adaptive images generated with \dalle (first row) for training the shape-adaptive generation model(SGM) and shape-adaptive VAE decoder(SVD). We show the SAM-based segmentation masks (left six columns) and the human-designed canvas masks (right two columns) for training SGM in the second row. The last row displays the augmented masks used as input conditions during SVD training, ensuring that the model learns to refine the augmented masks into the segmentation masks.}
\label{fig:dalle_training}
\end{figure}

\vspace{1mm}
\noindent\textbf{Shape-adaptive Image-Text Data Generation.}
To construct high-quality, shape-adaptive triplets for training our shape-adaptive generation models, precise alignment among the components is crucial.
Motivated by \dalle's exceptional ability to interpret and follow complex long prompts, along with its capability to produce high-quality and visually appealing images within a simple visual context surrounding a rectangular canvas, we have chosen \dalle as the engine for generating our training images. 

First, we use BLIP\cite{li2022blip} to generate detailed captions for LAION~\cite{schuhmann2022laion} images, creating a text prompt dataset that encompasses a broad spectrum of concepts. Second, we employ \dalle to create images based on these prompts, using the format ``\emph{Illustration of {prompt}. The whole scene is set against a clean white background, with no elements being cut off.}'' In this context, the {prompt} is crafted to direct the \dalle model to produce images that clearly differentiate the foreground canvas region from the background. Third, we use SAM~\cite{kirillov2023segment} to segment the foreground regions and generate irregular-shaped canvas masks and images accordingly, as depicted in Figure~\ref{fig:dalle_training}. For additional details on data processing, please see the supplementary. This process has resulted in approximately $80,000$ prompts, with each prompt yielding three unique images, culminating in a total of $240,000$ high-quality training instances.

\vspace{1mm}
\noindent\textbf{Shape-adaptive Attention.}
We use \(\mathbf{\Phi} \in \mathbb{R}^{c_{\rm{in}} \times h \times w}\) to represent the image latent features extracted by a VAE encoder before they are sent into the UNet of the diffusion model. We use \(\mathbf{\Phi}' \in \mathbb{R}^{n \times c}\) to represent the reshaped and transformed latent features that are sent into the multi-head cross-attention mechanism. By applying different linear projections, we transform \(\mathbf{\Phi}'\) into the query embedding space \(\mathbf{Q}\), and the text prompt embedding (or pixel embedding) into the key embedding space \(\mathbf{K}\) and value embedding space \(\mathbf{V}\) for cross-attention (or self-attention).
To accommodate our irregularly shaped canvas, we introduce a specialized variant: shape-adaptive attention scheme.

The key insight involves partitioning the entire image's feature maps into two groups: the foreground and the background. We use \(\mathbf{M}_A\) to denote the foreground pixels, the subscript \(fg\) to label the key and value embeddings associated with the regions inside the irregular canvas, and the subscript \(bg\) to label the key and value embeddings associated with the regions outside the irregular canvas.
The mathematical formulation is shown as follows:
\begin{equation}
    \begin{split}
        \textsf{ShapeAdaptive-MultiHeadAttention}(\mathbf{Q}, \mathbf{K}_{fg}, \mathbf{K}_{bg}, \mathbf{V}_{fg}, \mathbf{V}_{bg}) = \\
        \mathbf{M}_A \cdot \textsf{MultiHeadAttention}(\mathbf{Q}, \mathbf{K}_{fg}, \mathbf{V}_{fg}) \\
        + (1 - \mathbf{M}_A) \cdot \textsf{MultiHeadAttention}(\mathbf{Q}, \mathbf{K}_{bg}, \mathbf{K}_{bg}),
    \end{split}
\end{equation}
where we empirically discover that our shape-adaptive attention scheme can effectively minimize content creation outside the irregular canvas.

\vspace{1mm}
\noindent\textbf{Shape-adaptive Generation Model Training.}
Based on the above prepared $240,000$ shape-adaptive image-text pairs generated by \dalle and the proposed shape-adaptive attention scheme, we conduct the training of the shape-adaptive generation model following the controlnet-depth-sdxl-1.0~\cite{controlnet-depth-xl} by replacing the original depth map condition with the generated or hand-crafted canvas masks. During training, we fix the UNet part of the model and only fine-tune the ControlNet components.
We conduct the training on a cluster with $16\times$ A100 GPUs, set the batch size as $256$, and maintain a constant learning rate of 1e-6 throughout the training process, which spanned $60,000$ steps.

\vspace{1mm}
\noindent\textbf{\srm.}
Shape-adaptive generation model can generate user specified content within a designated area. However, there are a few drawbacks. First, there may still be solid color backgrounds and object shadows that interfere with the generation of the alpha mask (See $\bar{\mathbf{I}}$ in Figure~\ref{fig:overall_framework}). Second, the generated font effects are usually hard-edged which may not be preferred by the user.
To further improve the visual appealing of the content within the irregular canvas, suppress the undesired artifacts outside the canvas and offer a flexible control between readability and text-effect strength,
we propose to apply an additional shape-adaptive refinement model to refine the object's edges for a more natural appearance and generating a precise post-refinement alpha mask.

\vspace{1mm}
\noindent\textbf{Regeneration Strategy of Shape-adaptive Refinement Model.}
We first crop the output \(\Bar{\mathbf{I}}\) predicted by the shape-adaptive generation model following the font-shaped canvas mask \(\mathbf{M}\), and then paste the segmented font-shaped canvas region onto a rectangle white-board, resulting in  \(\Bar{\mathbf{I}}'\).
Next, we extract its latent representation \({\mathbf{z}_0}'\) and add noise to get \({\mathbf{z}_t}\) for \(t < T\).
We implement a regeneration strategy to start with \(\Bar{\mathbf{I}}'\) as the starting image and introducing a small amount of noise, disrupt the high-frequency signals while preserving the low-frequency components.
We find the diffusion model struggles to alter low-frequency signals during the denoising process, but concentrates on refining high-frequency elements to smooth out the object's edges.

Our shape-adaptive refinement model supports flexible control of readability and text-effect strength via noise strength. By using a larger noise strength value, the model tends to generate results with stronger text effect and vice versa. In our default setting, we set noise strength of shape-adaptive refinement model to 0.8. This value provides the model with enough flexibility to modify the character boundaries while ensuring the characters remain readable.

\begin{figure}[h]
\centering
\includegraphics[width=1.0\textwidth]{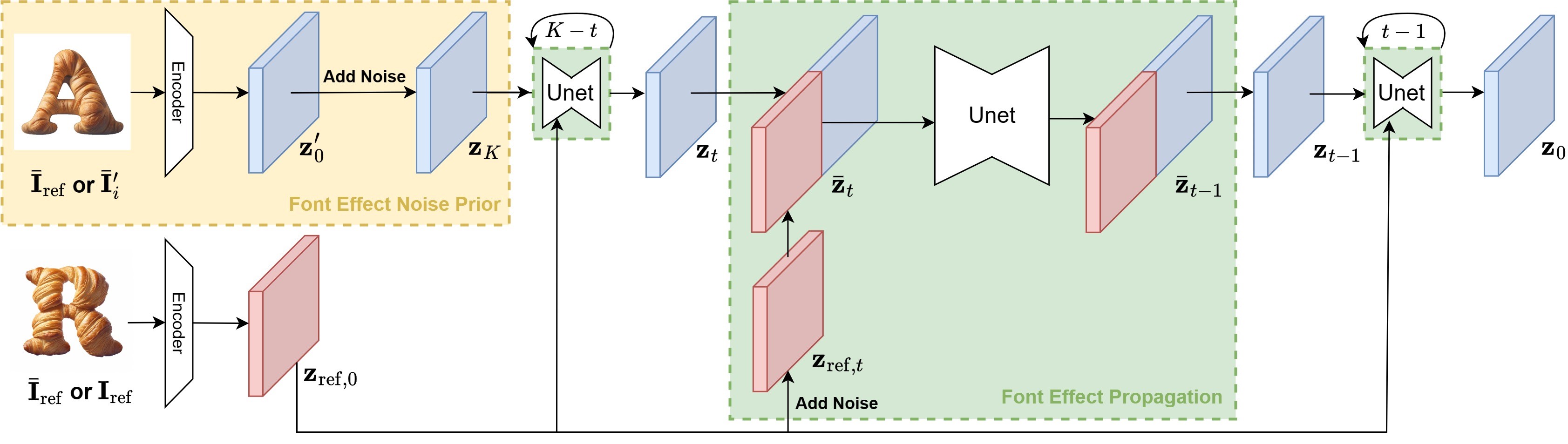}
\caption{Illustrating font effect noise prior and font effect propagation within shape-adaptive effect transfer (SAET) scheme. SAET can be applied on both shape-adaptive generation model (SGM) and shape-adaptive refinement model (SRM). When SAET is applied to SGM, we use $\bar{\mathbf{I}}_{\rm{ref}}$ for both font effect noise prior and font effect propagation. When SAET is applied to SRM (shown in figure), we use $\bar{\mathbf{I}}_{i}'$ for font effect noise prior and $\mathbf{I}_{\rm{ref}}$ for font effect propagation.}
\label{fig:font_effect_prop}
\end{figure}

\vspace{1mm}
\noindent\textbf{Shape-adaptive VAE Decoder (SVD).}
By applying a decoder to the denoised estimation \(\mathbf{z}_0\), we can generate a font effect image with refined edges, which may not confront to the given font-shaped canvas. This necessitates refining the alpha mask to enhance visual quality.
To this end, we propose fine-tuning a shape-adaptive VAE decoder capable of predicting an additional refined alpha mask associated with the decoded font effect image.
We simply augment the original VAE decoder with an additional input and output channel to facilitate mask conditioning and prediction.
During fine-tuning, we apply alpha mask augmentation to the original segmentation masks predicted with SAM~\cite{kirillov2023segment}, as shown in the third row of Figure~\ref{fig:dalle_training}.
In summary, the fine-tuned VAE decoder is capable of predicting a refined alpha mask in addition to decoding the image.

\subsection{Shape-adaptive Effect Transfer}
As we have ensured the creation of high-quality visual content on any given irregular font-shaped canvas, the next critical challenge is ensuring a consistent font effect across multiple characters. We propose a shape-adaptive effect transfer (SAET) scheme to transfer the reference font effect from one image to all target letter font images. 
SAET can be applied to any diffusion-like models. The key idea involves modulating the inputs and outputs of the diffusion model as well as influencing the latent feature of the denoising process, denoted as \(\mathbf{z}_t\). In our case, we applied SAET to shape-adaptive diffusion model including both SGM and SRM. Therefore, we can reformulate the original Equation~\ref{eq:effect_transfer} as follows:
\begin{align}
\mathbf{I}_i, \mathbf{M}_{\mathbf{I}_i} &= \bar{h}(
\mathbf{M}_i\;|\;\mathbf{T},\; \mathbf{M}_{\rm{ref}}, \;\mathbf{I}_{\rm{ref}},\;\mathbf{M}_{\mathbf{I}_{\rm{ref}}}
),\; \rm{i}\in\{1, \cdots, n\},
\end{align}
In the following, we differentiate the style source (reference image) from the style recipient (target image) for clarity.

\vspace{1mm}
\noindent\textbf{Framework Overview.}
The efficacy of the shape-adaptive effect transfer scheme is attributed to two pivotal factors: first, it provides the target image with an effect prior based on the reference image; second, it iteratively integrates effect information from the reference image throughout the denoising process, resulting in a target image with a consistent font effect.
Figure~\ref{fig:overall_framework} also illustrates the overall framework of our shape-adaptive effect transfer approach.

\vspace{1mm}
\noindent\textbf{Font Effect Noise Prior.} 
Drawing inspiration from SDEdit~\cite{meng2021sdedit}, we devise a font effect noise prior scheme by initializing target font images with partially noised latents derived from the original reference font effect image. This approach enhances the model's ability to generate styles consistently. The overall implementation is depicted in Figure~\ref{fig:font_effect_prop}.

\vspace{1mm}
\noindent\textbf{Font Effect Propagation.}
We further propose to propagate the font effect information from the reference font image to the target font image following:
at any denoising stage \(t\) within a UNet, given the target image's latent \(\mathbf{z}_{t}\) and the reference image's latent \(\mathbf{z}_{\mathrm{ref},0}\), we escalate \(\mathbf{z}_{\mathrm{ref},0}\) to the same noise level as \(\mathbf{z}_{t}\), yielding \(\mathbf{z}_{\mathrm{ref},t}\). We then concatenate \(\mathbf{z}_{t}\) with \(\mathbf{z}_{\mathrm{ref},t}\) to obtain \(\bar{\mathbf{z}}_t=\mathrm{Concat}(\mathbf{z}_{\mathrm{ref},t}, \mathbf{z}_{t})\), which is then processed through UNet for denoising. After deducing the noise component, we selectively utilize the noise pertaining to \(\mathbf{z}_{t}\) for denoising \(\mathbf{z}_{t}\) to achieve \(\mathbf{z}_{t-1}\), iterating this step until reaching \(\mathbf{z}_{0}\).
The effect propagation between the source latents and the target latents mainly happen within the self-attention modules.
Figure~\ref{fig:font_effect_prop} illustrates the detailed process.

We modify both the shape-adaptive generation model and shape-adaptive refinement model to support processing the concatenated latent representations of a source font effect image and a target font image with font effect prior.
We empirically find setting the noise strengths with different values within SGM and SRM achieves the best results. Refer to the supplementary for more details.

\vspace{1mm}
\noindent\textbf{Discussion.}
Our empirical findings suggest that our method is resilient to variations in the reference font shape, yielding consistent results across a wide range of reference font shapes. We have observed that choosing a reference character with a larger foreground area is beneficial. This is interpreted as the larger foreground providing more informative units for the self-attention mechanism, thereby enhancing the generation of new characters. In practice, we often use the letter `R' from the specified font as our reference for generation due to its typically large foreground area. Refer to supplementary material for more details. Additionally, our approach demonstrates flexibility across different language scripts, having been successfully applied to fonts in Chinese, Japanese, and Korean in our extended experiments.

\section{Experiments}
\subsection{\ourbenchmark benchmark}
We introduce the \ourbenchmark benchmark, which comprises 145 test cases, to enable comprehensive comparisons. These prompts vary in length and are categorized into five themes: Nature, Material, Food, Animal, and Landscape. The character sets extend beyond English, incorporating Chinese, Japanese, and Korean characters, offering a diverse linguistic and cultural representation. This benchmark serves as the foundation for all data analyses and comparative studies conducted in this work. For detailed information on its construction, please refer to the supplementary material.

\begin{table}[ht]
\begin{minipage}{0.6\linewidth}
\centering
\caption{Ablation results of SGM}
\label{tab:model_ablation}
\tablestyle{1pt}{1.1}
\resizebox{1\linewidth}{!}
{
\begin{tabular}{l|cc}
Model & \maskclip↑ & \maskclipout ↓\\ 
\shline
SDXL-ControlNet-Canny &  26.03 & 21.52 \\
SDXL-ControlNet-Depth &  24.11 & 23.24 \\
SGM trained w. Est-depth & 24.51 & 18.28\\
SGM trained w. Cropped Est-depth & 24.11 & 18.22 \\
SGM w.o SAA & 27.10 & 22.07 \\
SGM & \textbf{27.26} & \textbf{18.11}\\
\end{tabular}
}
\end{minipage}
\hfill
\begin{minipage}{0.35\textwidth}
\begin{minipage}{1\textwidth}
    \centering
    \caption{Shape-Adaptive Effect Transfer vs. StyleAligned}
    \vspace{-3mm}
    \label{tab:effect_abaltion}
    \tablestyle{8pt}{1.}
    \resizebox{1\linewidth}{!}
{
    \begin{tabular}{l|c|c}
    Model &  CLIP-I↑ & DINO↑ \\
    \shline
    \ourname w.o. SAET & 81.02 & 54.27 \\
    \ourname w. StyleAligned & 82.77 & 60.79\\
    \ourname & \textbf{84.63} & \textbf{67.07}\\
    \end{tabular}
    }
\end{minipage}
\begin{minipage}{1\textwidth}
    \centering
    \caption{Comparison with Adobe Firefly}
    \vspace{-3mm}
    \tablestyle{11pt}{1.}
    \resizebox{1\linewidth}{!}
{
    \begin{tabular}{l|cc}
    Model & CLIP↑ & CLIP-I↑\\
    \shline
    Firefly & 28.48 & 81.74\\
    \ourname & \textbf{29.44} & \textbf{84.63}\\ 
    \end{tabular}
    }
    \label{tab:adobe_qualitative}
    \end{minipage}
    \end{minipage}
\end{table}

\subsection{Ablation Study on \model}
To assess the ability of models to accurately generate content within the font canvas area in accordance with provided prompts, we introduced the \maskclip and \maskclipout metrics. These metrics make use of an additional mask to direct the evaluation towards the intended areas, both inside and outside the canvas. In the calculations for \maskclip and \maskclipout, we mask areas outside the canvas in white and subsequently average these altered CLIP similarity scores across the benchmark.

\vspace{1mm}
\noindent\textbf{Comparison between \sgm and Rectangle-canvas based Diffusion Models.}
Figure \ref{fig:rectangle_canva_results} showcases the qualitative results from conventional diffusion models trained for rectangle canvas. SDXL faces challenges in performing the font effect generation task due to missing shape-specific guidance. Conversely, SDXL-Inpaint, while not tailored to fill the entire area with designated content, often produces barely recognizable shapes. Both SDXL-ControlNet-Canny and SDXL-ControlNet-Depth are capable of processing masked inputs; however, their training primarily focuses on matching prompts with the entire rectangle image canvas, inadvertently causing prompt content to appear outside the intended shape area. This misalignment adversely affects their \maskclipout scores, as detailed in Table \ref{tab:model_ablation}. Additionally, the lack of targeted control guidance within the shape leads to diminished \maskclip scores for these models. We also note that it is impractical to apply SDXL-ControlNet-Segmentation to our task. The reason is that ControlNet requires a precomputed segmentation map of finite number of classes, and it is hard to estimate a reasonable segmentation map that fits the irregular font shape while following the complex semantics of user prompts.

\vspace{1mm}
\noindent\textbf{Training Objective.}
We fine-tuned two depth models using our image dataset: the first model employed estimated depth maps, while the second utilized depth maps that were cropped according to the shape mask. Table \ref{tab:model_ablation} shows that both models underperformed in all metrics, underscoring the difficulty depth models face in generating content within specified areas, despite being trained with our data. However, our training approach significantly increases the models' flexibility, a key factor in the superior performance of our model.

\begin{figure}[htbp]
\tiny
\centering
\begin{minipage}{0.3\textwidth}
\vspace{-4mm}
\begin{tabular}{ccc}
\tiny \raisebox{.1\textwidth}{\rotatebox[origin=l]{90}{w.o SAA}} &
\includegraphics[width=0.45\textwidth]{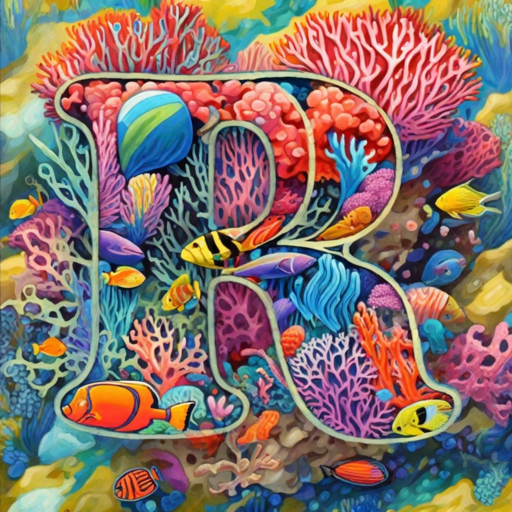}&
\includegraphics[width=0.45\textwidth]{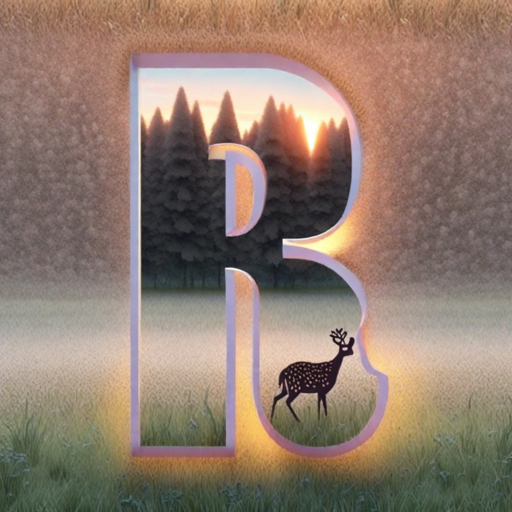}\\
\tiny \raisebox{.1\textwidth}{\rotatebox[origin=l]{90}{w. SAA}} &
\includegraphics[width=0.45\textwidth]{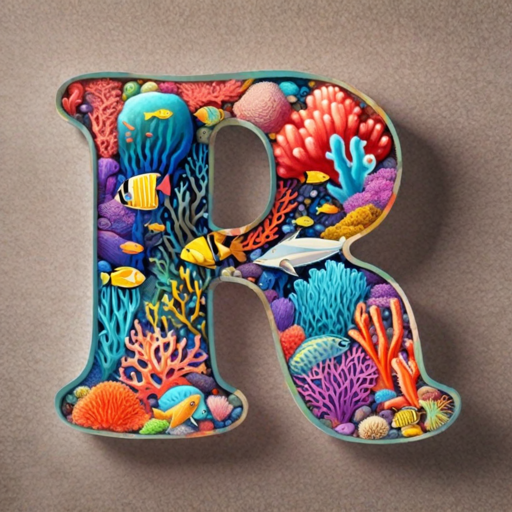}&
\includegraphics[width=0.45\textwidth]{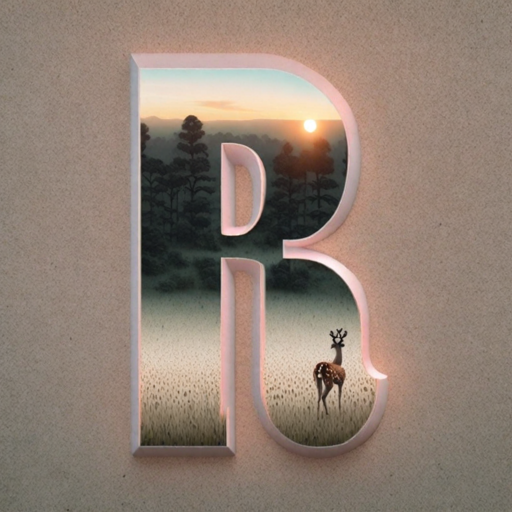}
\end{tabular}
\vspace{-3mm}
\caption{\small Effect of Shape-adaptive Attention.}
\label{fig:ablation_mixattention}
\end{minipage}
\begin{minipage}{0.6\textwidth}
\vspace{-3mm}
\begin{tabular}{ccccc}
\tiny \raisebox{.07\textwidth}{\rotatebox[origin=l]{90}{SAM}} &
\includegraphics[width=0.225\textwidth]{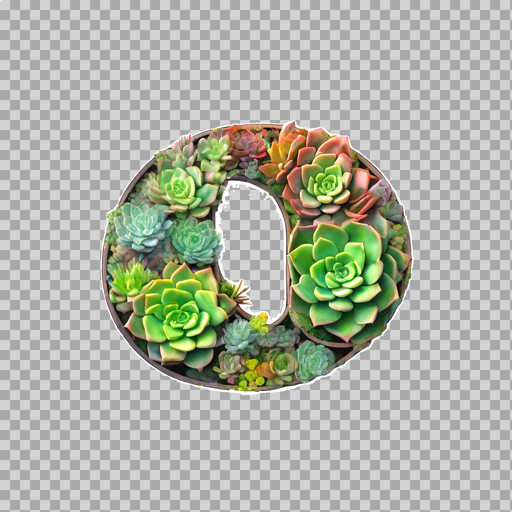}&
\includegraphics[width=0.225\textwidth]{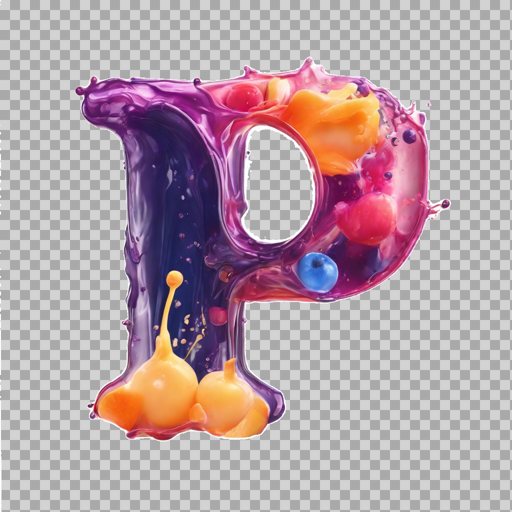}&
\includegraphics[width=0.225\textwidth]{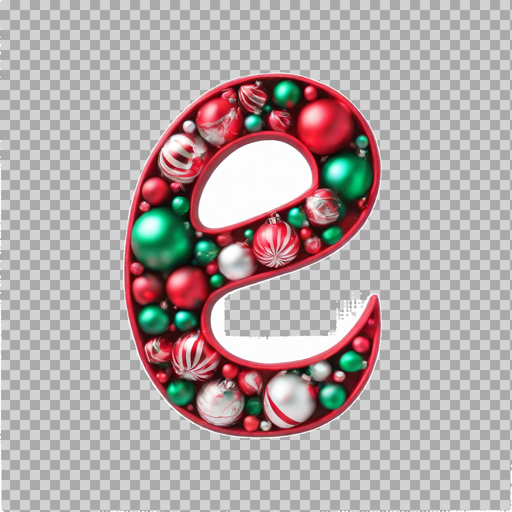}&
\includegraphics[width=0.225\textwidth]{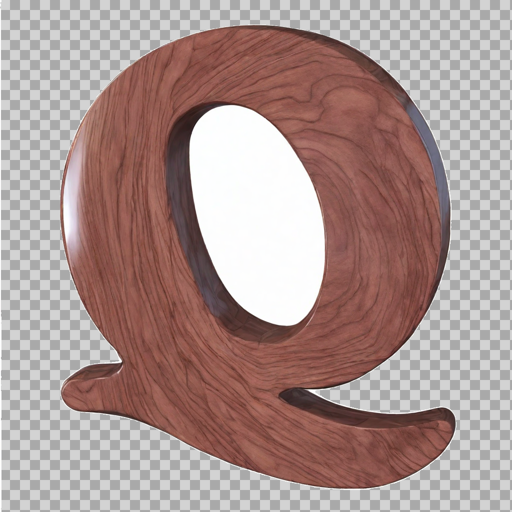}\\
\tiny \raisebox{.07\textwidth}{\rotatebox[origin=l]{90}{SVD}} &
\includegraphics[width=0.225\textwidth]{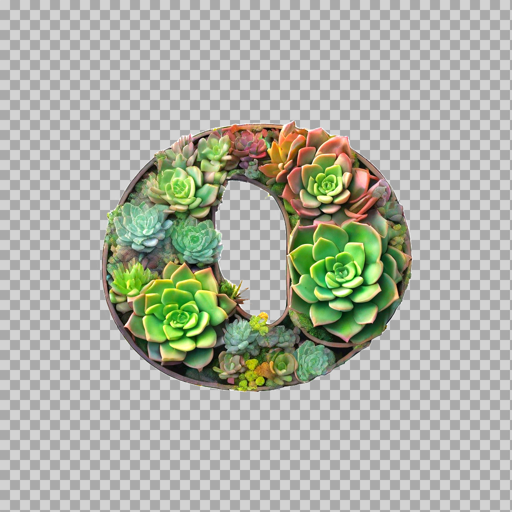}&
\includegraphics[width=0.225\textwidth]{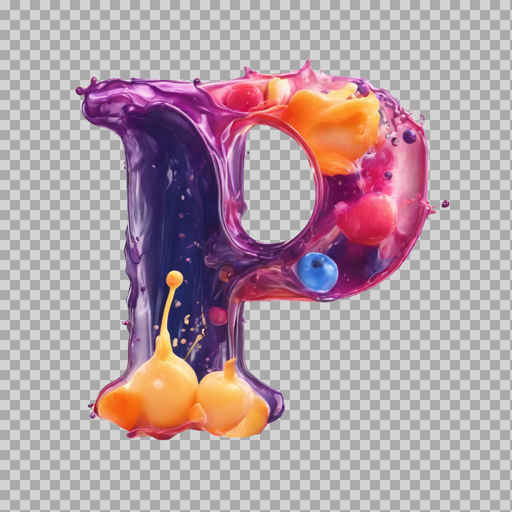}&
\includegraphics[width=0.225\textwidth]{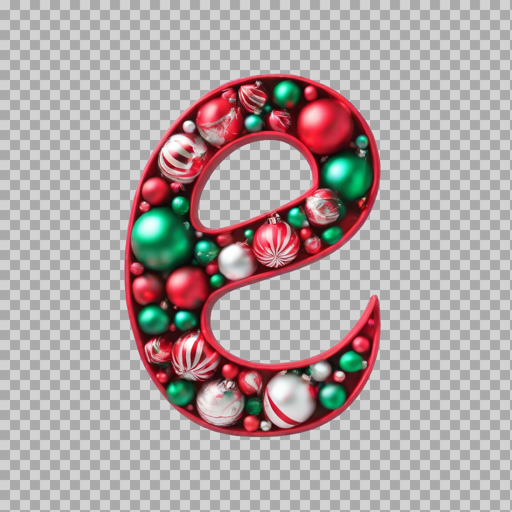}&
\includegraphics[width=0.225\textwidth]{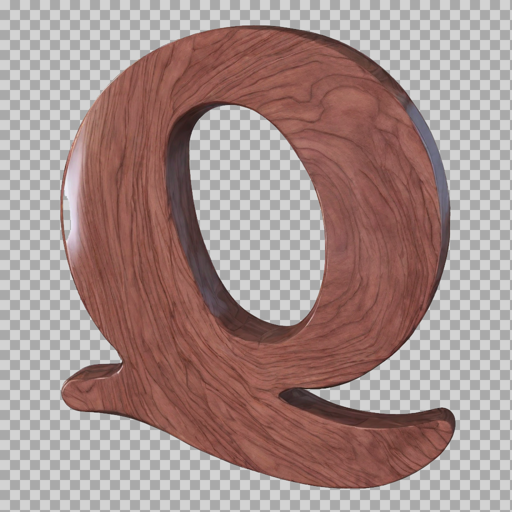}
\end{tabular}
\caption{\small Effect of Shape-adaptive VAE Decoder.}
\label{fig:ablation_SVD}
\end{minipage}
\end{figure}

\vspace{1mm}
\noindent\textbf{Ablation on Shape-adaptive Attention.}
As detailed in Table \ref{tab:model_ablation} and illustrated in Figure \ref{fig:ablation_mixattention},  our shape-adaptive attention can not only markedly reduce the generation of background elements but also enhances the creation and intricacy of foreground content.

\begin{figure}[htbp]
\tiny
\centering
\begin{minipage}{1\textwidth}
\centering
\tiny
\begin{tabular}{lcc}
\rotatebox[origin=l]{90}{w.o SAET}  & \includegraphics[width=0.45\textwidth]{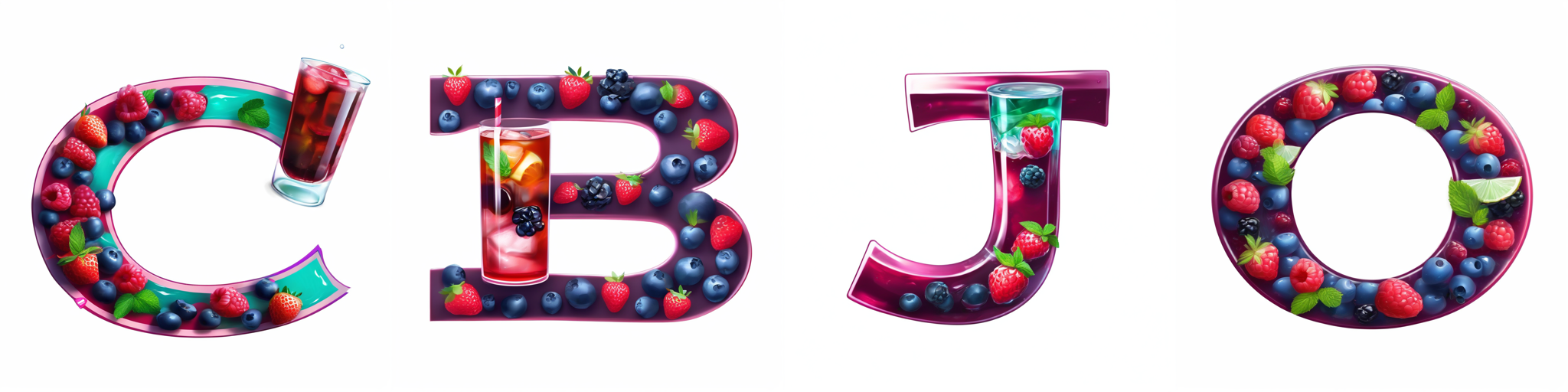} & \includegraphics[width=0.45\textwidth]{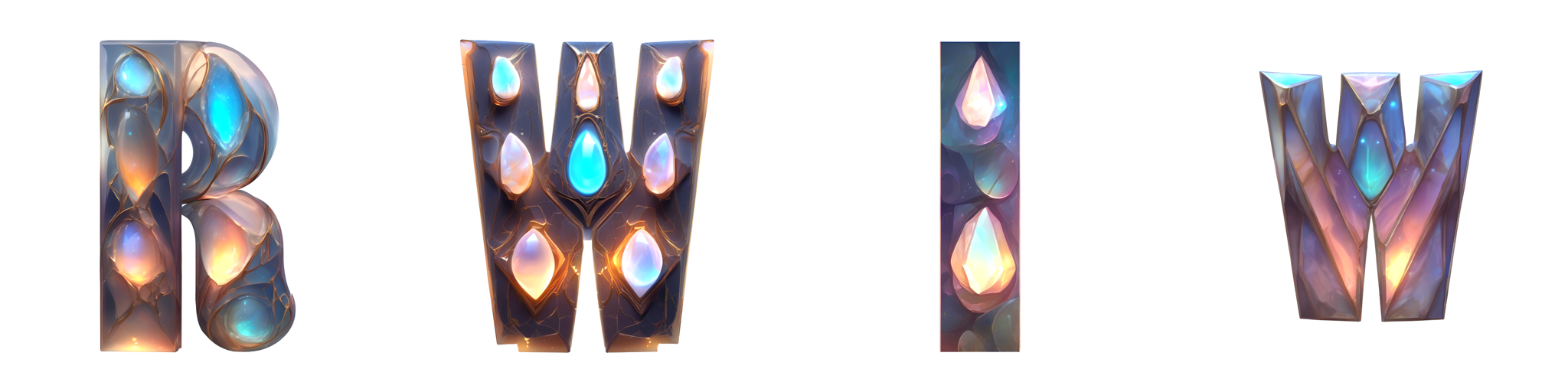}  \\
\rotatebox[origin=l]{90}{StyleAligned}  & \includegraphics[width=0.45\textwidth]{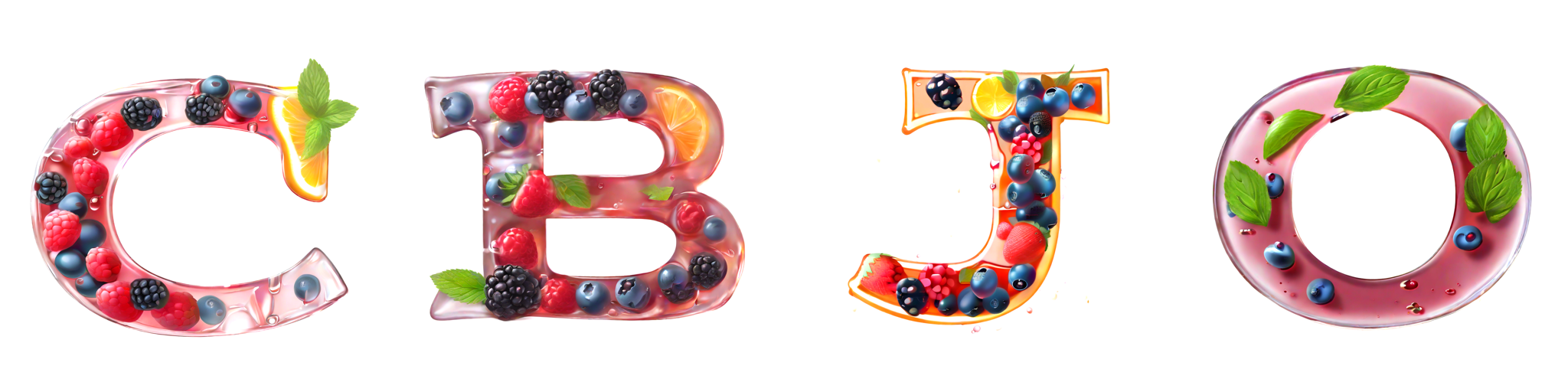} & \includegraphics[width=0.45\textwidth]{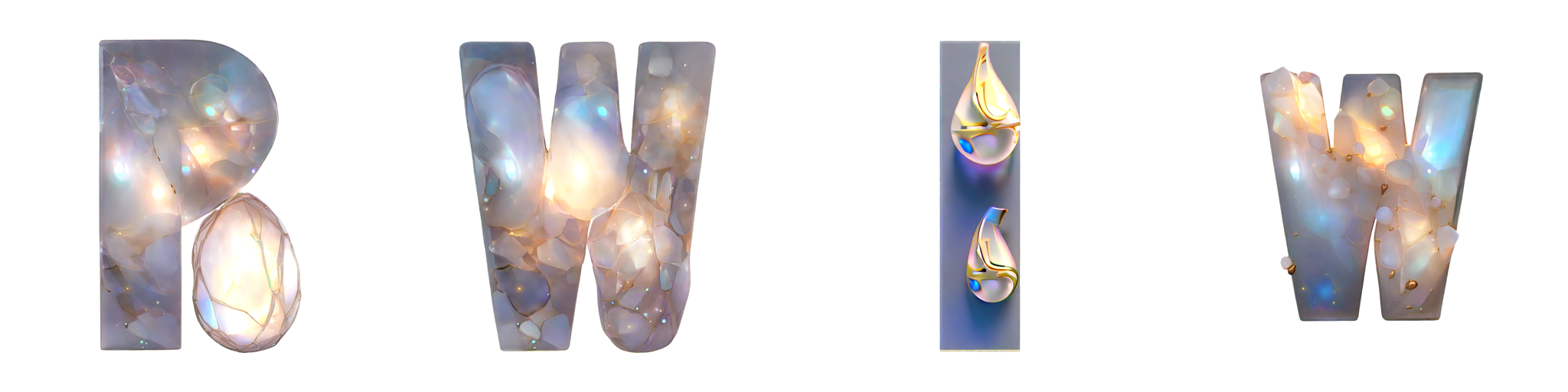}  \\
\raisebox{.01\textwidth}{\rotatebox[origin=l]{90}{\ourname}}  & \includegraphics[width=0.45\textwidth]{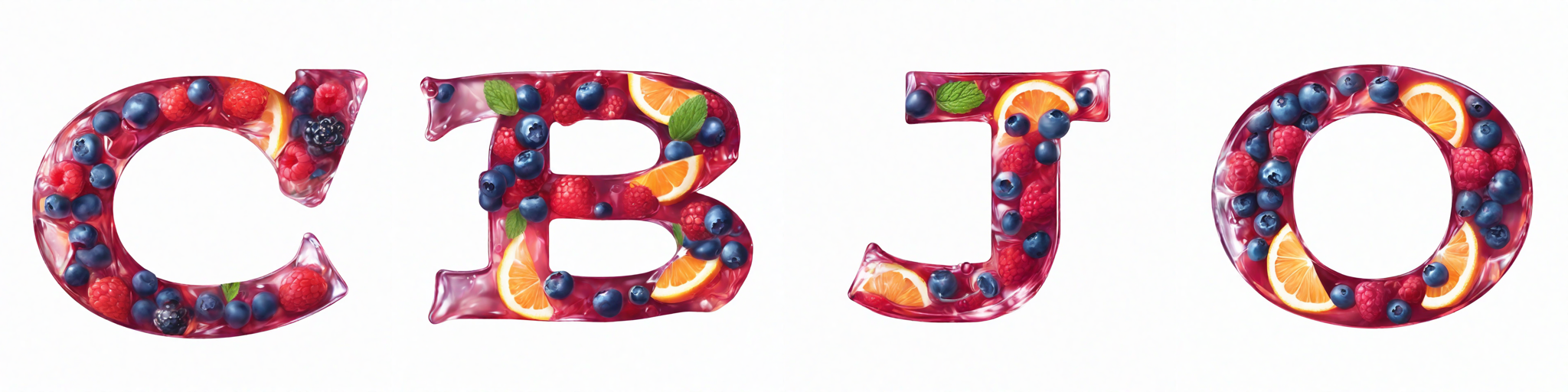} & \includegraphics[width=0.45\textwidth]{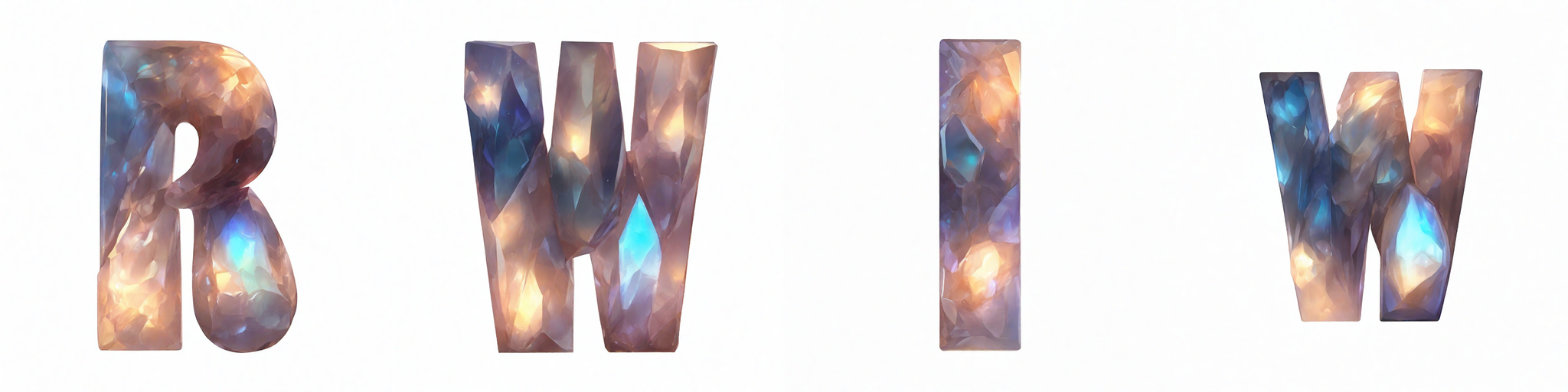}  \\
\end{tabular}
\caption{\ourname vs. StyleAligned: qualitative comparison results.}
\label{fig:effect_abaltion}
\end{minipage}
\begin{minipage}{1\textwidth}
\centering
\includegraphics[width=0.975\textwidth]{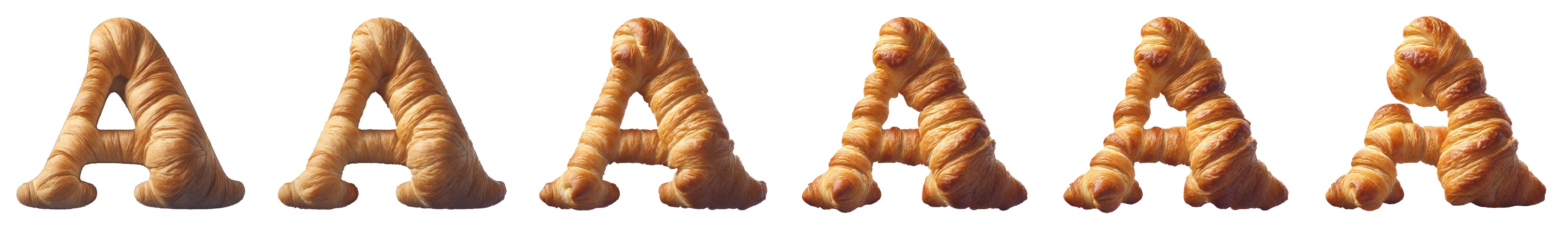}
\begin{tabular}{m{1.95cm} m{1.95cm} m{1.95cm} m{1.75cm} m{1.95cm} m{0.5cm}}
\ \ \ 0.0&\  0.4 &\ 0.6 & 0.8 & 0.85 &\ 0.9 \\
\end{tabular}
\caption{Shape-adaptive refinement model results with different noise strength.}
\label{fig:different_noise_strength}
\end{minipage}
\begin{minipage}{1\textwidth}
\begin{minipage}{0.45\textwidth}
\centering
\includegraphics[width=0.9\textwidth]{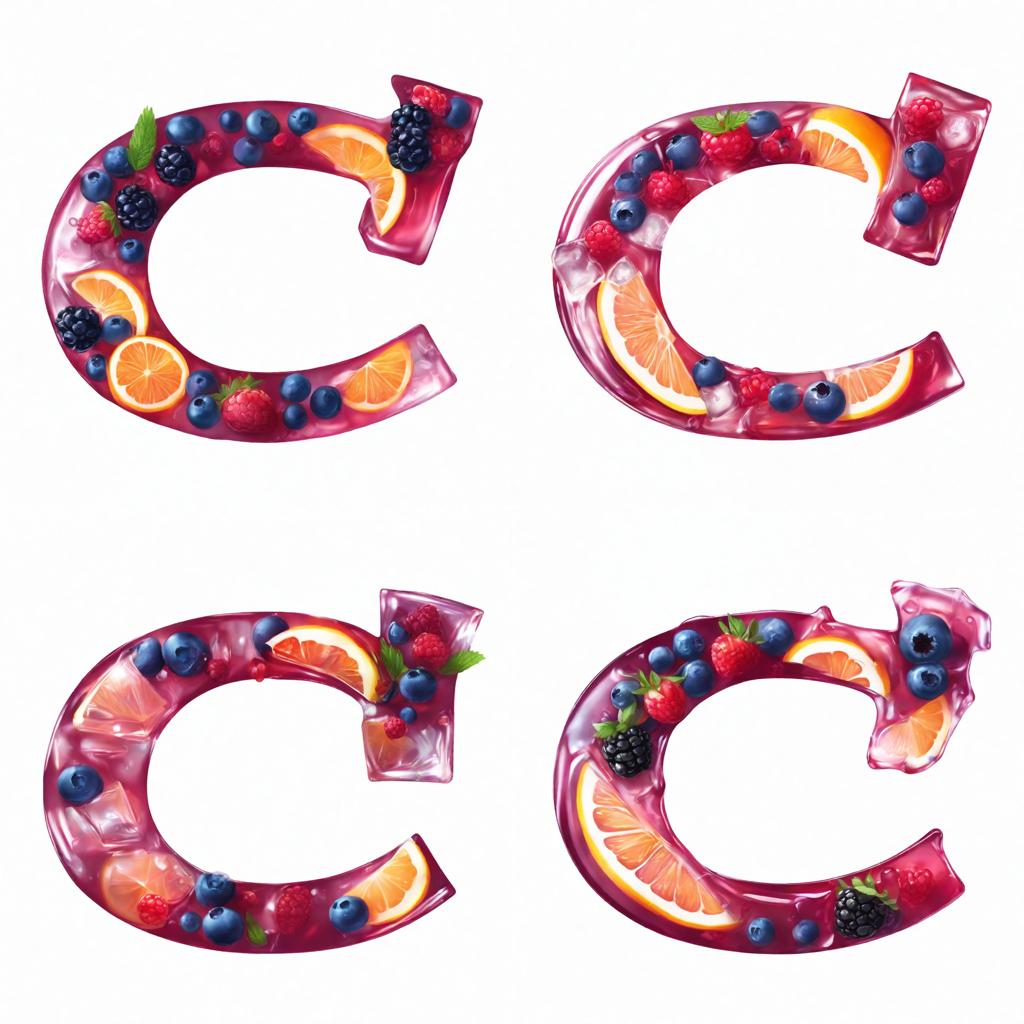}
\caption{Font-effect variation with different seed.}
\label{fig:effect_abaltion_different}
\end{minipage}
\hspace{3mm}
\begin{minipage}{0.55\textwidth}
\begin{tabular}{
C{0.1cm} C{0.22\textwidth} C{0.22\textwidth} C{0.22\textwidth} C{0.22\textwidth}}
& stones & cherry blossom &  neons & bananas
\end{tabular}
\begin{tabular}{lcccccccc}
\raisebox{.1\textwidth}{\rotatebox[origin=c]{90}{Shape}}  & \includegraphics[width=0.22\textwidth]{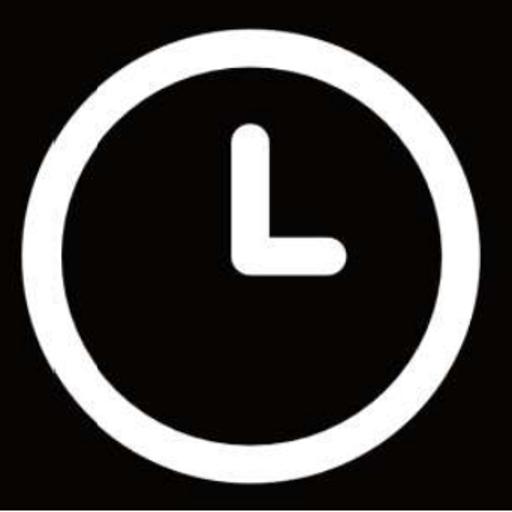} &
\includegraphics[width=0.22\textwidth]{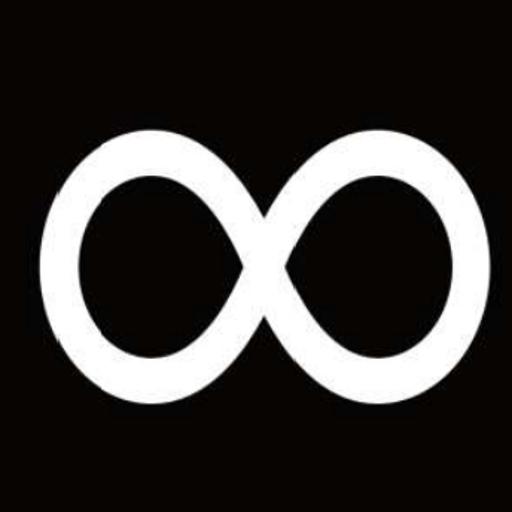} &\includegraphics[width=0.22\textwidth]{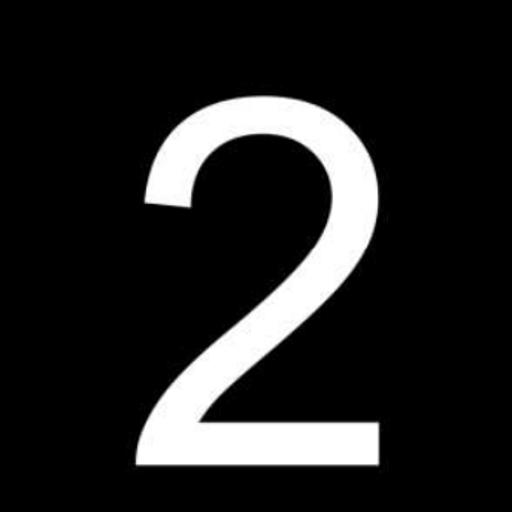}& \includegraphics[width=0.22\textwidth]{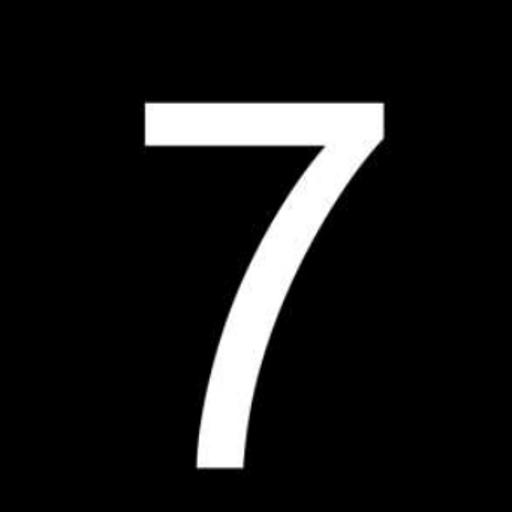}  \\
\raisebox{.11\textwidth}{\rotatebox[origin=c]{90}{A2G}} & \includegraphics[width=0.22\textwidth]{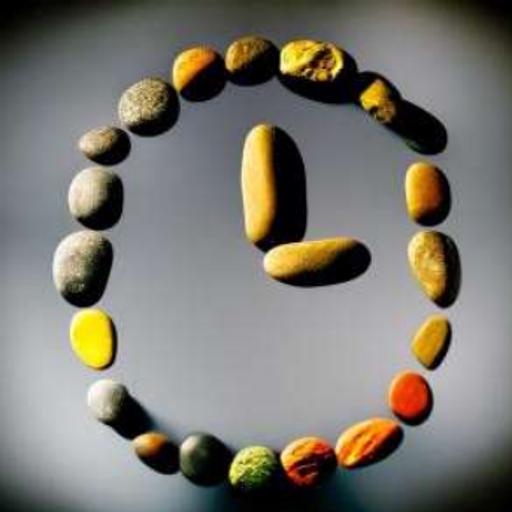} & \includegraphics[width=0.22\textwidth]{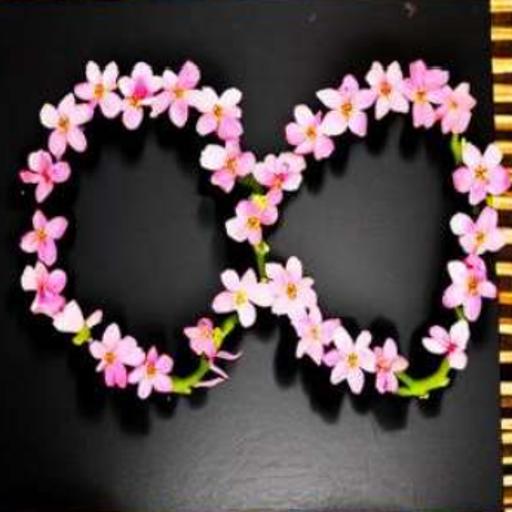}& \includegraphics[width=0.22\textwidth]{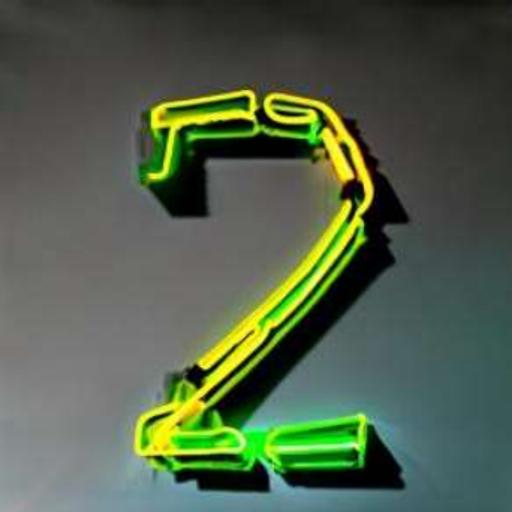}& \includegraphics[width=0.22\textwidth]{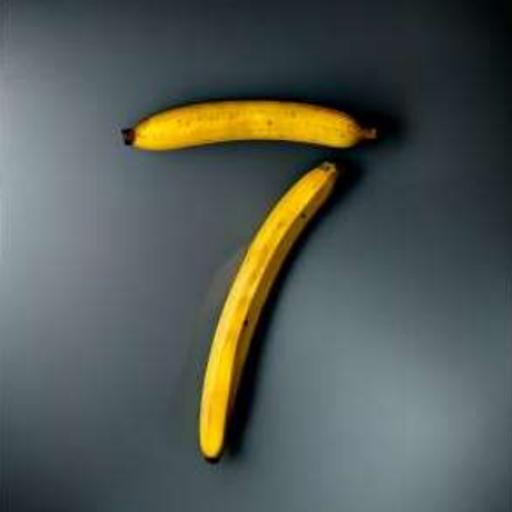}  \\
\raisebox{.12\textwidth}{\rotatebox[origin=c]{90}{\ourname}} & \includegraphics[width=0.22\textwidth]{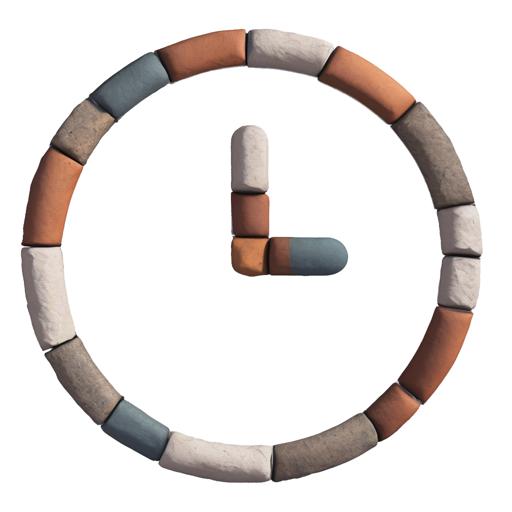} & \includegraphics[width=0.22\textwidth]{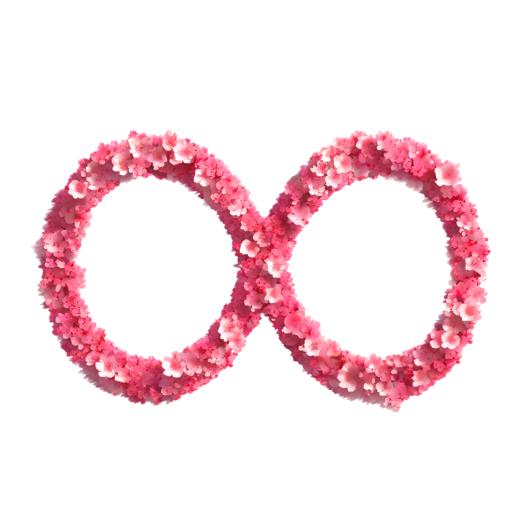}& \includegraphics[width=0.22\textwidth]{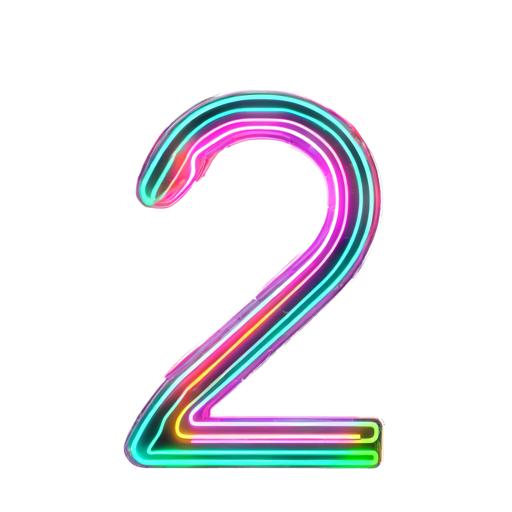}
& \includegraphics[width=0.22\textwidth]{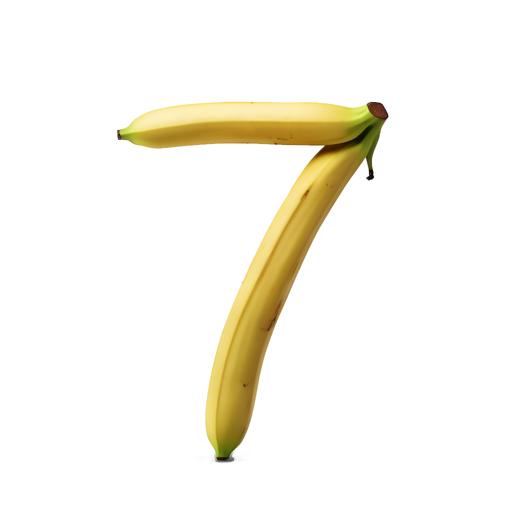} \\
\end{tabular}
\vspace{-3mm}
\caption{Comparison with Anything to Glyph(A2G).}
\label{fig:a2g_comparaison}
\end{minipage}
\end{minipage}
\begin{minipage}{1\textwidth}
\tiny
\centering
\vspace{2mm}
\begin{tabular}{
C{2.2cm} C{2.2cm} C{2.2cm} C{2.2cm} C{2.2cm}}
... drawing ... scculent garden & metal &  red and green holiday ornaments &  room &  ... fish ... coral reef ... fauvist painting ...
\end{tabular}
\begin{tabular}{c c}
  \raisebox{.135\textwidth}{\rotatebox[origin=c]{90}{\ourname   \ \ \ \ \  Adobe Firefly}}   & \includegraphics[width=0.95\textwidth]{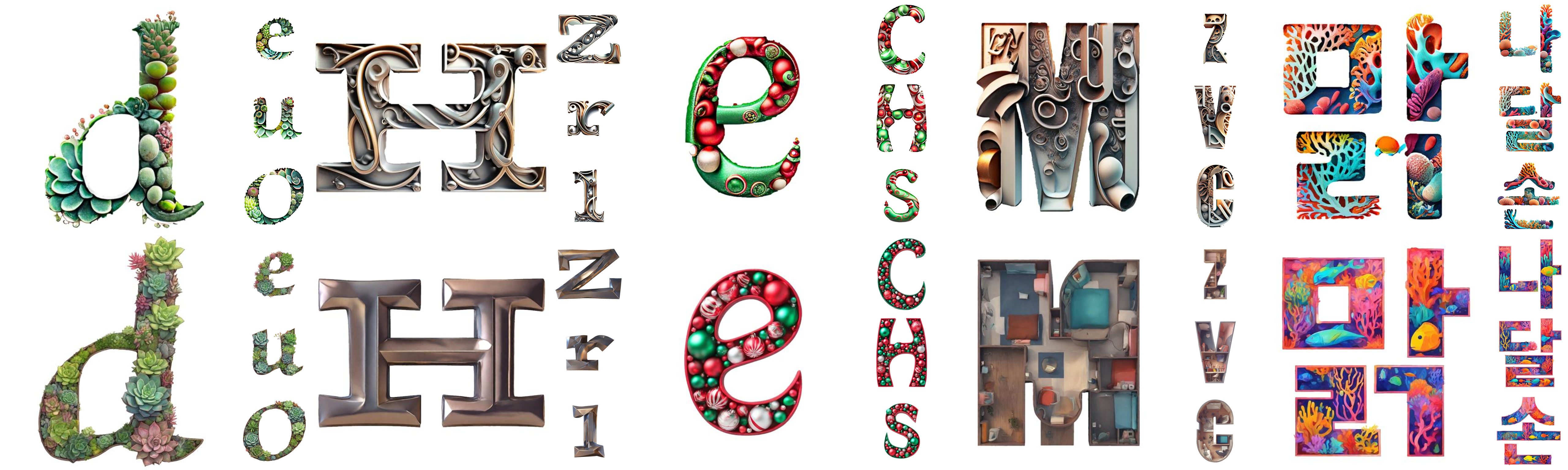}
\end{tabular}
\caption{\small Qualitative comparison with Adobe Firefly Text Effect.}
\label{fig:Firefly}
\end{minipage}
\end{figure}

\vspace{1mm}
\noindent\textbf{Ablation on Noise Strength of \srm.}
Figure~\ref{fig:different_noise_strength} demonstrates that as the noise strength increases, the boundaries of character `A' become more flexible. Our default setting of 0.8 strikes an ideal balance between readability and text effect strength. However, we have also noticed that in some cases, even with a higher noise strength, the model still tends to strictly follow the original character shapes.
For more discussion about this phenomenon, please refer to the supplementary material.

\vspace{1mm}
\noindent\textbf{Shape-adaptive VAE Decoder.}
Our comparison between SVD and SAM, depicted in Figure \ref{fig:ablation_SVD}, reveals that SAM tends to generate masks that are somewhat coarse, occasionally leaving blank spaces within characters uncleaned. SVD, however, leverages the input mask as guidance, significantly lowering the likelihood of errors and producing more accurate alpha masks.

\begin{figure}[htbp]
\tiny
\centering
\begin{minipage}{1\textwidth}
\tiny
\centering
\includegraphics[width=0.9\textwidth]{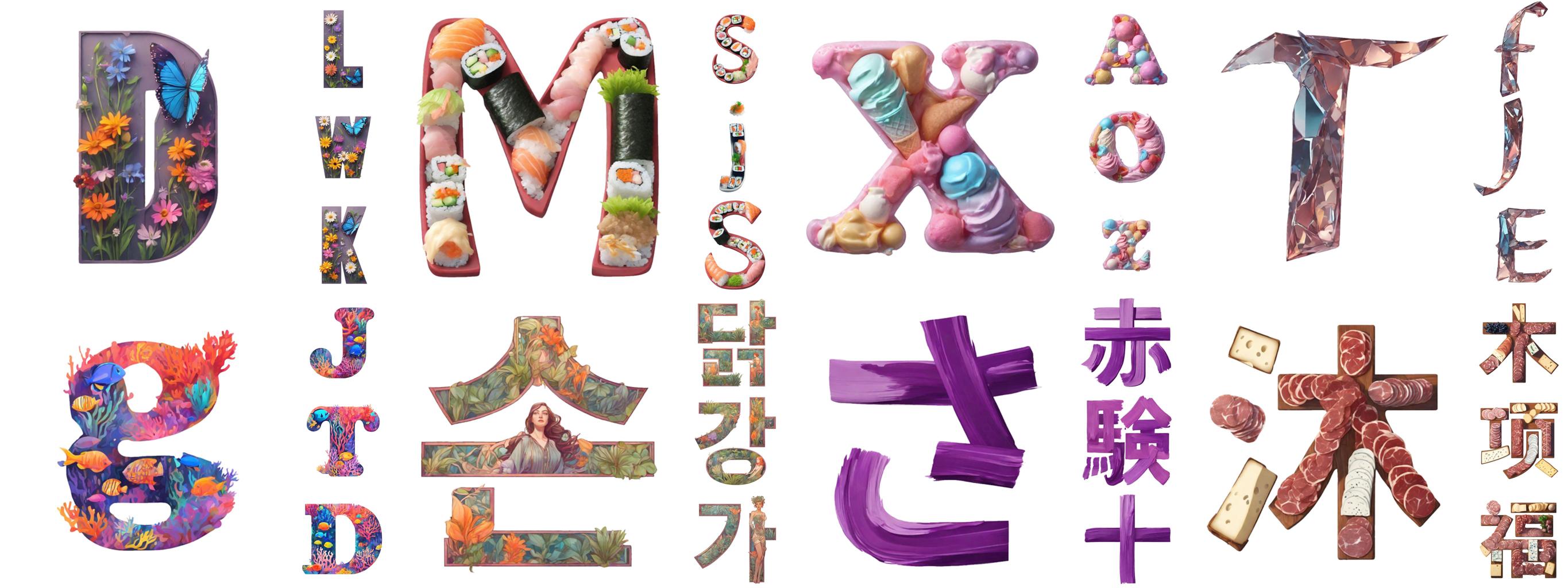}
\includegraphics[width=0.9\textwidth]{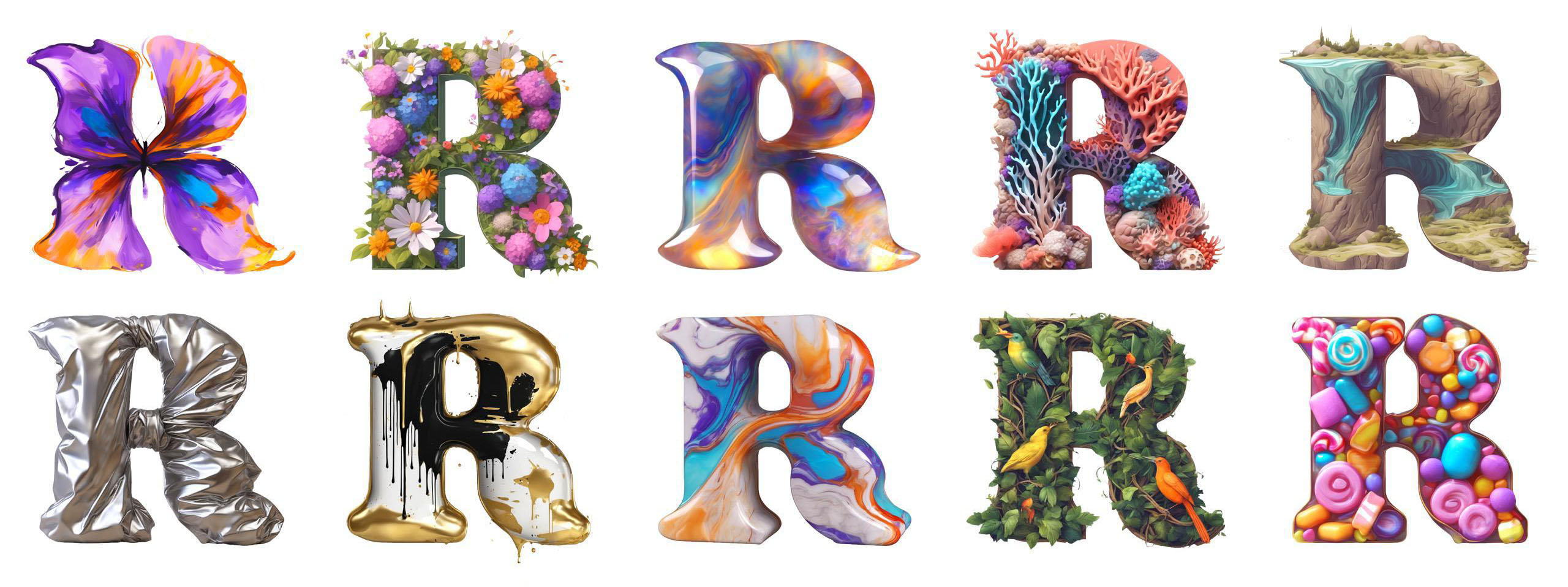}
\vspace{-4mm}
\caption{\small Qualitative font-effect results generated with our \ourname.}
\label{fig:Showcaes}
\end{minipage}
\end{figure}

\subsection{Ablation Study on Shape-adaptive Effect Transfer}

In this section, we employ the CLIP-I score and DINO score to assess the visual font effect similarity across the generated characters as in \cite{ruiz2023dreambooth}.

\vspace{1mm}
\noindent\textbf{Comparison with Baseline and StyleAligned~\cite{hertz2023style}.}
Our baseline for comparison involves the shape-adaptive diffusion model without SAET, where each character is generated independently using uniform seed. We also substitute StyleAligned for our SAET to evaluate its performance. The outcomes, illustrated in Table \ref{tab:effect_abaltion}, reveal that models utilizing SAET significantly outperform those that do not in terms of both CLIP-I and DINO score. Figure \ref{fig:effect_abaltion} highlights that, despite a fixed generation seed, maintaining style consistency across different shapes proves challenging for models without SAET. Conversely, SAET enables the generation of visually similar yet uniquely detailed outcomes by varying seeds (Figure \ref{fig:effect_abaltion_different}), which is advantageous for real-world applications by preventing the replication of identical results for repeated characters.

\subsection{Comparison with State-of-the-Art}
\noindent\textbf{Comparison with Anything to Glyph.}
Given the non-availability of Anything to Glyph we selectively compare images from this method for our analysis. As illustrated in Figure \ref{fig:a2g_comparaison}, across all test instances, our \ourname consistently produces the content dictated by the prompt while effectively preserving the input shape's integrity. Moreover, the font effect generated by our approach is created without an accompanying background, simplifying its integration into user-specific designs. Aesthetically, our designs boast a more cohesive color palette and are free from incongruous shadows on the letter forms. We note that Figure \ref{fig:a2g_comparaison} also demonstrates that our model can accept inputs of any shape, not just limited to characters.

\noindent\textbf{Comparison with Adobe Firefly.}
Figure \ref{fig:Firefly} shows outputs from both frameworks. Firefly's outputs feature high contrast and a consistent style but often include mismatched patterns, reducing character clarity and aesthetic value. In contrast, \ourname presents outputs with cohesive colors, diverse styles, and clear linework, enhancing letter integration. For shape fidelity, both frameworks maintain character legibility, though Firefly's can appear fragmented with missing strokes due to its aesthetic issues. Stylistically, both are largely consistent, though Firefly occasionally shows minor discrepancies. Regarding prompt fidelity, both generally follow prompt instructions, but Firefly struggles more with style-related prompts. Table \ref{tab:adobe_qualitative} delineates the quantitative comparison on \ourbenchmark benchmark between our results and those by Firefly, focusing on the CLIP Score and CLIP-I Score, reflective of Prompt Fidelity and Style Consistency, respectively. Our analysis underscores our methodology's superior performance over Firefly across these metrics. Moreover, we provides more visualization results in Figure~\ref{fig:Showcaes}.

\noindent\textbf{User Study and GPT-4V evaluation.}
We engaged 25 evaluators, including 10 professionals, to assess the benchmark results, and similar assessments were conducted for GPT-4V. Participants rated the outcomes using four metrics to determine which were superior. The findings, displayed in Figure \ref{fig:user_study}, confirm the superiority of our \ourname over Adobe Firefly in every category. We also have similar results for GPT-4V with a 65\% win rate in aesthetics, 76\% in shape fidelity and 74\% in style consistency. Refer to the supplementary for more details.

\section{Conclusion}
We introduced \ourname, an innovative system crafted for generating coherent and consistent visual content specifically designed for font shapes. The system consists of two principal components: a shape-adaptive diffusion model that tackles the challenge of creating content on irregular canvases, and a shape-adaptive effect transfer scheme ensuring uniformity across characters. Furthermore, we present the \ourbenchmark benchmark, a tool developed for the quantitative evaluation of our method's efficacy. Our empirical studies demonstrate that \ourname adeptly responds to user prompts, creating high-quality and aesthetically pleasing font effects. Notably, it surpasses both previous studies and the commercial solution Adobe Firefly in all metrics assessed.

\bibliographystyle{splncs04}
\bibliography{egbib}

\newpage
\section{Supplementary Material}

\begin{CJK*}{UTF8}{gbsn}

\subsection{Detailed Font Effect Prompts in the Figures of the Main Paper}

In this section, we elaborate on the prompts associated with figures from the main paper, as detailed across Tables \ref{tab:fig1_font_effect}, \ref{tab:fig2_font_effect}, \ref{tab:fig7_font_effect}, \ref{tab:fig8_font_effect}, \ref{tab:fig9_font_effect}, \ref{tab:fig12_font_effect}, and \ref{tab:fig13_font_effect}.

\subsection{More Details about Shape-adaptive Image-Text Data Generation}

\subsubsection{Irregular Canvas Mask Generation}
For image generation, we derive irregular canvas masks using the SAM segmentation model or manually designed templates. SAM effectively segments areas with uniform background colors when provided with a full-image bounding box prompt, allowing for precise canvas mask predictions by inverting the initial segmentation masks. To augment the diversity and enhance the quality of our training masks, we introduced custom masks in the shapes of rectangles or ellipses, featuring variable aspect ratios. We compute the aspect ratio, denoted as ${r}$, using the following formula:
\begin{equation}
    {r} = \frac{\min(1, 1-0.3 \times (0.5-X))}{\min(1, 1-0.3 \times (X-0.5))}
\end{equation}
where $X \sim \texttt{Beta}(\alpha=1.5, \beta=1.5)$. Images are cropped to fit these masks and placed on a white background, with random resizing applied. The resize scale ${s}$ is defined by:
\begin{equation}
    {s} = 1 - 0.4 \times Y
\end{equation}
where $Y \sim \texttt{Beta}(\alpha=5, \beta=5)$.

\subsubsection{Mask Augmentation for Shape-adaptive VAE Decoder Training}
To improve the accuracy of mask prediction by the shape-adaptive VAE decoder, we employed a training strategy that utilizes masks which are marginally expanded or contracted. Within the confines of the original mask's bounding box, the augmented mask undergoes modification through the application of Gaussian noise. This process involves mapping the intermediate values to 255 to achieve expansion, or to 0 for contraction. During training, we apply an additional MSE loss between the ground truth mask and the predicted mask. We conduct the training on $4\times$ A100 GPUs with a batch size of 32 and maintained a constant learning rate of 1e-6 throughout the training process, which spanned 80,000 steps.

\begin{table}[htbp]
\begin{minipage}[t]{1\linewidth}  
\centering  
\tablestyle{2pt}{1.1}  
\resizebox{1.0\linewidth}{!}  
{  
\begin{tabular}{l|l|>{\centering\arraybackslash}m{15cm}}  
Character(s) & Font Type & Prompt \\  
\shline
Font & COOPBL &\small{croissant} \\ \hline
Studio & COOPBL & \small{a cozy cottage with smoke} \\ \hline
Fo & COOPBL & \small{a bustling city square alive with the sound of vendors} \\ \hline
nt & COOPBL & \small{flower lei} \\ \hline 
Stu & COOPBL & \small{jungle vine and bird} \\ \hline
dio & COOPBL & \small{a coral reef, alive with color and bustling marine life, captured in the vivid colors of a fauvist painting} \\ \hline
F & COOPBL & \small{the majestic sight of a waterfall cascading down rocky terrains, enveloped in a misty spray, rendered in the rich, textured layers of an oil painting} \\ \hline
o & COOPBL & \small{a detailed drawing of a succulent garden, showcasing various textures and shades of green, with tiny flowers emerging} \\ \hline 
n & COOPBL & \small{impressionist landscape of a japanese garden in autumn, with a bridge over a koi pond} \\ \hline 
t & COOPBL & \small{art nouveau painting of a female botanist surrounded by exotic plants in a greenhouse} \\ \hline
S & COOPBL & \small{bundle of colorful electric wires} \\ \hline
t & COOPBL & \small{an old bridge arching over a serene river} \\ \hline
u & COOPBL & \small{a charcuterie board, featuring thinly sliced prosciutto, salami, and a variety of aged cheeses} \\ \hline
d & COOPBL & \small{sushi} \\ \hline
i & COOPBL & \small{driftwood} \\ \hline
o & COOPBL & \small{a night sky ablaze with stars} \\
\end{tabular}  
}
\caption{  
\footnotesize{Illustrating the font effect prompts listed in Figure 1 are arranged in a sequence that progresses from the top left to the bottom right.}}
\label{tab:fig1_font_effect}
\end{minipage}
\end{table}

\subsection{More Implementation Details}
In \ourname, both the input and output images are standardized at a resolution of $1024 \times 1024$. The prompts for UNet models are formulated using the template ``a shape fully made of \{prompt\}, artistic, trending on artstation.'' with a fixed denoising step count of 50. Classifier-free guidance (CFG) is implemented with a guidance scale of 6.0. All results are inferenced in fp16 mode.

For the ControlNet components in both the shape-adaptive generation model and its shape-adaptive effect transfer variant, the control scale is uniformly set at 1.0.
Denoising in the shape-adaptive generation model begins with pure noise. The noise strength for the shape-adaptive refinement model is adjusted to 0.8. For both the shape-adaptive generative model with shape-adaptive effect transfer and its refinement version, the noise strengths are set at 0.9 and 0.8, respectively.

Both the encoder \cite{madebyollin/sdxl-vae-fp16-fix} and the UNet \cite{stable-diffusion-xl-base-1.0} are shared and frozen for shape-adaptive generative model and shape-adaptive refinement model.

\subsection{More Details of \ourbenchmark Benchmark}

Our benchmark comprises three distinct elements: prompts, font types, and characters, designed to rigorously evaluate the generative capabilities of models. The prompts are organized into five principal themes—Nature, Material, Food, Animal, and Landscape—to cover a broad spectrum of visual textures and compositional challenges. Specifically, Nature, Material, and Food themes are associated with textures that can be uniformly applied across different characters, while Animal and Landscape themes are intended to test the model’s ability to handle complex compositions.

To further assess the model's interpretive skills, prompts are also categorized by their length (short, medium, long), introducing varying levels of complexity. This results in a total of 100 prompts, either inspired by Adobe Firefly examples or generated by GPT-4V.

In terms of font types and characters, our selection aims to challenge the model across a variety of shapes, focusing on different stroke thicknesses and aspect ratios. The benchmark includes five English font types: COOPBL, SANVITO, POSTINO, HOBO, and POPLAR. Additionally, a single font type, SourceHansSansHeavy, is chosen for each of the Chinese, Japanese, and Korean languages to ensure representation of diverse script systems.

Each test involves four unique characters to check style consistency. The English tests incorporate 52 capital and lowercase letters, whereas for Chinese, Japanese, and Korean, 10 characters of varying complexity are selected from each language to represent complexity diversity.

The benchmark comprises 145 sets of prompts, fonts, and characters, with 100 for English corresponding to all prompts, and 15 each for Chinese, Japanese, and Korean, using three from each category to cover all prompt lengths. The full benchmark is recorded in Tables \ref{tab:bench_en1}, \ref{tab:bench_en2}, \ref{tab:bench_cn}, \ref{tab:bench_jp} and \ref{tab:bench_kr}.

\subsection{Details about GPT-4V evaluation}

The prompt template used for GPT-4V evaluation is shown as following:

\begin{pkprompt}
You have been enlisted as an expert designer to evaluate the outputs of two font effect generation tools. These tools are designed to embed specified prompt content into four designated characters, and their goal is to create images that are legible, aesthetically appealing, and stylistically coherent.

The performance of these font effect generators is mixed, with some outputs being superior to others. The order of these two images are randomized. Your role involves using a professional and impartial lens to compare the results from both generators based on the given prompts and four letters, across four distinct metrics.

Your evaluation criteria include:

Aesthetics: Assess the visual appeal of the generated characters, considering the composition's harmony and the attractiveness of the color scheme. A well-designed output should leverage the text's shape to organize visual elements effectively, featuring rich colors that are well-proportioned and theme-appropriate, with balanced lighting and contrast. In contrast, inferior designs may present visual elements that appear abruptly segmented or distorted, utilize colors that clash with the theme, or include elements that create visual discomfort due to extreme brightness, darkness, or saturation.

Font Shape Fidelity: Evaluate the text's legibility. The characters' outlines may either remain unaltered or be adapted creatively based on the visual theme. A quality design retains the characters' original outlines or modifies them in a way that enhances readability, tailoring the boundaries to fit the visual theme. Conversely, a subpar design might lead to character confusion due to compromised design elements.

Font Style Consistency: Judge the uniformity of design style among characters, including aspects like color use, design motifs, and brushwork. An outstanding design achieves consistency while allowing for unique adjustments to each character's shape, preventing the design from becoming repetitive. On the other hand, a flawed design is evident when design elements visibly clash or create a sense of discord.

Prompt Fidelity: Ascertain if each character faithfully adheres to the prompt. This involves checking if every aspect of the prompt, such as objects, adjectives, and the overall design style, is accurately depicted. Ideally, the design should fully reflect the prompt's elements rather than partially or tangentially. While minor additions for design enhancement are acceptable, the prompt's components should remain central to the design. Superior designs will seamlessly and comprehensively incorporate all elements from the prompt, whereas deficient designs lack or alter essential elements, making it challenging to connect the visual output back to the original theme.

Kindly structure your feedback in JSON format, specifying your preference (Image1, Image2, or Draw) using keyword "Preference" for each metric, along with your reasoning with keyword "Reason". Please make your reason concise and each reason shall be less than 40 words.

For this test, two design images are generated using prompt \{prompt\} and the characters \{characters\}.
\end{pkprompt}

\subsection{More Comparison Results with Adobe Firefly}

\begin{figure}[t] 
\tiny
\centering
\vspace{2mm}
\begin{tabular}{
C{0.3cm} C{2.7cm} C{2.7cm} C{2.7cm} C{2.7cm}}
& a red fox prowling through a snowy forest & a butterfly flitting among wildflowers &  juice splash &  ... pancakes ... maple syrup ... blueberries
\end{tabular}
\begin{tabular}{c c}
  \raisebox{.175\textwidth}{\rotatebox[origin=c]{90}{\ourname   \ \ \ \ \ \ \ \ \ \  Adobe Firefly}}   & \includegraphics[width=0.95\textwidth]{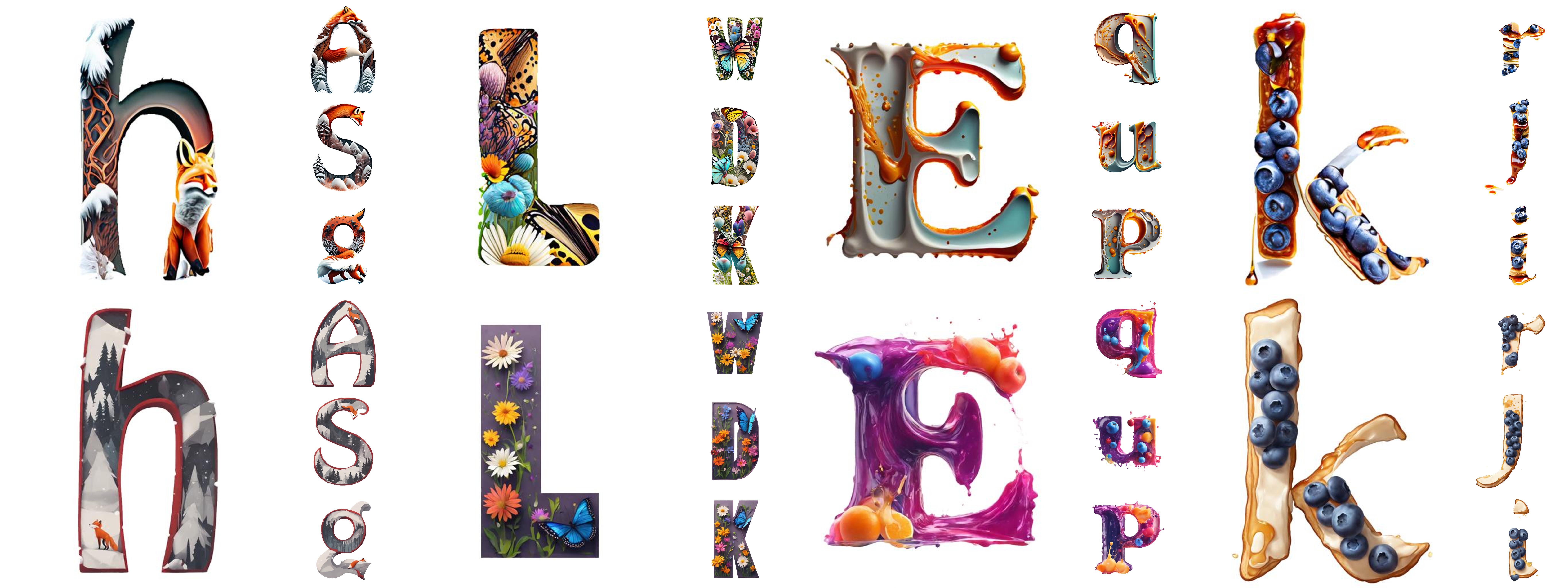}
\end{tabular}
\begin{tabular}{
C{0.1cm} C{2.7cm} C{2.7cm} C{2.7cm} C{2.7cm}}
& leafy pothos & .. waterfall .. rocky terrains ... misty spray ... oil painting &  decay &  a charcuterie board ... sliced prosciutto, salami ... cheeses
\end{tabular}
\begin{tabular}{c c}
  \raisebox{.175\textwidth}{\rotatebox[origin=c]{90}{\ourname   \ \ \ \ \ \ \ \ \ \  Adobe Firefly}}   & \includegraphics[width=0.95\textwidth]{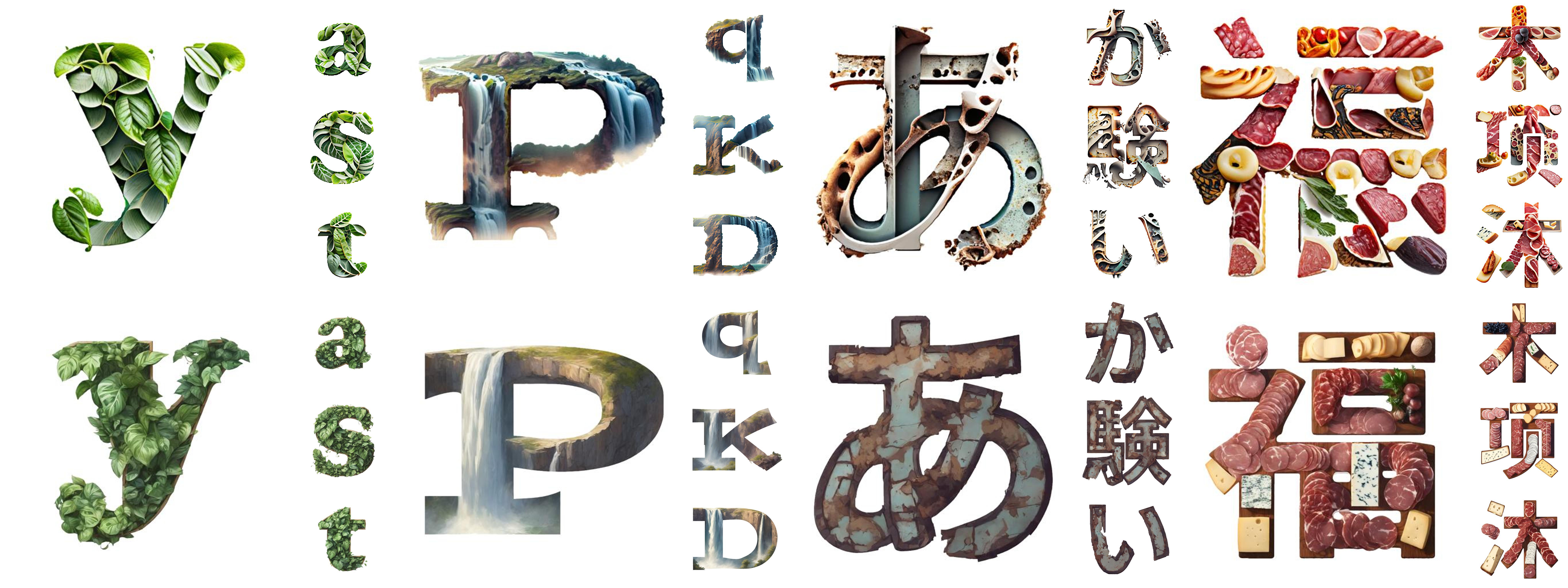}
\end{tabular}
\caption{\small More Qualitative Comparisons with Adobe Firefly Text Effect.}
\label{fig:MoreFirefly}
\end{figure}

This section presents further comparison results, as illustrated in Figure \ref{fig:MoreFirefly}, with the corresponding prompts detailed in Table \ref{tab:prompt_firefly}.

\begin{table}[htbp]
\begin{minipage}[t]{1\linewidth}  
\centering  
\tablestyle{2pt}{1.1}  
\resizebox{1.0\linewidth}{!}  
{  
\begin{tabular}{l|l|>{\centering\arraybackslash}m{15cm}}  
Character & Font type & Prompt \\  
\shline
R & COOPBL & \small{ice cream} \\ \hline 
A & COOPBL & \small{jungle vine and bird} \\
\end{tabular}  
}
\caption{  
\footnotesize{Illustrating the font effect prompts listed in Figure 2 are arranged in a sequence that progresses from the top to buttom.}}
\label{tab:fig2_font_effect}
\end{minipage}
\begin{minipage}[t]{1\linewidth}  
\centering  
\tablestyle{2pt}{1.1}  
\resizebox{1.0\linewidth}{!}  
{  
\begin{tabular}{l|l|>{\centering\arraybackslash}m{15cm}}  
Character & Font Type & Prompt \\  
\shline
R & COOPBL & \small{a coral reef, alive with color and bustling marine life, captured in the vivid colors of a fauvist painting} \\ \hline 
R & POPLAR & \small{a spotted deer grazing in a meadow at dawn} \\
\end{tabular}  
}
\caption{  
\footnotesize{Illustrating the font effect prompts listed in Figure 7 are arranged in a sequence that progresses from the left to right.}}
\label{tab:fig7_font_effect}
\end{minipage}
\begin{minipage}[t]{1\linewidth}  
\centering  
\tablestyle{2pt}{1.1}  
\resizebox{1.0\linewidth}{!}  
{  
\begin{tabular}{l|l|>{\centering\arraybackslash}m{15cm}}  
Character & Font Type & Prompt \\  
\shline
o & COOPBL & \small{a detailed drawing of a succulent garden, showcasing various textures and shades of green, with tiny flowers emerging} \\ \hline 
P & COOPBL & \small{juice splash} \\ \hline
e & HOBO & \small{red and green holiday ornaments} \\ \hline
Q & COOPBL & \small{glossy cherry wood, its surface smooth and reflecting a warm, deep sheen} \\
\end{tabular}  
}
\caption{  
\footnotesize{Illustrating the font effect prompts listed in Figure 8 are arranged in a sequence that progresses from the left to right.}}
\label{tab:fig8_font_effect}
\end{minipage}
\begin{minipage}[t]{1\linewidth}  
\centering  
\tablestyle{2pt}{1.1}  
\resizebox{1.0\linewidth}{!}  
{  
\begin{tabular}{l|l|>{\centering\arraybackslash}m{15cm}}  
Characters & Font Type & Prompt \\  
\shline
CBJO & POSTINO & \small{a berry-infused iced tea, sweetened just right and served with ice, garnished with fresh berries and a sprig of mint for a refreshing summer quencher} \\ \hline
RWIw & POPLAR & \small{luminous moonstones, their surfaces alive with an ethereal, shifting glow} \\
\end{tabular}  
}
\caption{  
\footnotesize{Illustrating the font effect prompts listed in Figure 9 are arranged in a sequence that progresses from the left to right.}}
\label{tab:fig9_font_effect}
\end{minipage}
\begin{minipage}[t]{1\linewidth}  
\centering  
\tablestyle{2pt}{1.1}  
\resizebox{1.0\linewidth}{!}  
{  
\begin{tabular}{l|l|>{\centering\arraybackslash}m{15cm}}  
Characters & Font Type & Prompt \\  
\shline
deuO & SANVITO & \small{a detailed drawing of a succulent garden, showcasing various textures and shades of green, with tiny flowers emerging} \\ \hline
ZrHl & POSTINO & \small{metallic} \\ \hline
eCHs & HOBO & \small{red and green holiday ornaments} \\ \hline
zMVC & POPLAR & \small{room} \\ \hline
\begin{CJK}{UTF8}{mj}맑나달손\end{CJK}
 & SourceHanSansKRHeavy & \small{a coral reef, alive with color and bustling marine life, captured in the vivid colors of a fauvist painting} \\
\end{tabular}  
}
\caption{  
\footnotesize{Illustrating the font effect prompts listed in Figure 12 are arranged in a sequence that progresses from the left to right.}}
\label{tab:fig12_font_effect}
\end{minipage}
\begin{minipage}[t]{1\linewidth}  
\centering  
\tablestyle{2pt}{1.1}  
\resizebox{1.0\linewidth}{!}  
{  
\begin{tabular}{l|l|>{\centering\arraybackslash}m{15cm}}  
Character(s) & Font Type & Prompt \\  
\shline
LwDK & POPLAR & \small{a butterfly flitting among wildflowers} \\ \hline
sMjS & HOBO & \small{sushi} \\ \hline
AXOz & COOPBL & \small{ice cream} \\ \hline
fJTE & SANVITO & \small{broken glass} \\ \hline
gJTD & COOPBL & \small{a coral reef, alive with color and bustling marine life, captured in the vivid colors of a fauvist painting} \\ \hline
\begin{CJK}{UTF8}{mj}닭강가손\end{CJK} & SourceHanSansKRHeavy & \small{art nouveau painting of a female botanist surrounded by exotic plants in a greenhouse} \\ \hline
\begin{CJK}{UTF8}{min}赤さ験十\end{CJK} & SourceHanSansJPHeavy & \small{purple paint brush stroke} \\ \hline
木项福沐 & SourceHanSansCNHeavy & \small{a charcuterie board, featuring thinly sliced prosciutto, salami, and a variety of aged cheeses} \\ \hline
R & COOPBL & \small{a vibrant butterfly captured in the lively, spontaneous strokes of an expressionist painting} \\ \hline
R & COOPBL & \small{a garden bursting with blooms in the spring sunshine} \\ \hline
R & COOPBL & \small{polished opals, displaying a mesmerizing play of colors within their depths} \\ \hline
R & COOPBL & \small{coral reef} \\ \hline
R & COOPBL & \small{rivers} \\ \hline
R & COOPBL & \small{plastic wrap} \\ \hline
R & COOPBL & \small{black and gold dripping paint
} \\ \hline
R & COOPBL & \small{color marble} \\ \hline
R & COOPBL & \small{jungle vine and bird} \\ \hline
R & COOPBL & \small{candy} \\ 
\end{tabular}  
}
\caption{  
\footnotesize{Illustrating the font effect prompts listed in Figure 13 are arranged in a sequence that progresses from the top left to bottom right.}}
\label{tab:fig13_font_effect}
\end{minipage}
\end{table}

\begin{table}[htbp]
\vspace{-3mm}
\begin{minipage}[t]{1\linewidth}  
\centering  
\tablestyle{2pt}{1.1}  
\resizebox{1.0\linewidth}{!}  
{  
\begin{tabular}{l|l|l|>{\centering\arraybackslash}m{15cm}}  
Characters & Font Type & Category & Prompt \\  
\shline
jWNF & POPLAR & Animal & \small{dragon} \\ \hline
BYPf & POPLAR & Animal & \small{peacock} \\ \hline
IUma & HOBO & Animal &\small{panda} \\ \hline
WoMZ & HOBO & Animal &\small{puppy} \\ \hline
JAxd & POSTINO & Animal &\small{kitties} \\ \hline
LpXn & SANVITO & Animal &\small{girrafe} \\ \hline
srdc & SANVITO & Animal &\small{pegasus} \\ \hline
RzHE & COOPBL & Animal &\small{snake} \\ \hline
tHFk & POSTINO & Animal &\small{firefly} \\ \hline
Gilq & POSTINO & Animal &\small{colourful starfish} \\ \hline
AhSg & HOBO & Animal &\small{a red fox prowling through a snowy forest} \\ \hline
QPaG & COOPBL & Animal &\small{a camel silhouetted against a desert sunset} \\ \hline
ovSY & POPLAR & Animal &\small{a spotted deer grazing in a meadow at dawn} \\ \hline
KyDu & SANVITO & Animal &\small{a school of fish swimming in a coral reef} \\ \hline
LwDK & POPLAR & Animal &\small{a butterfly flitting among wildflowers} \\ \hline
nrMZ & HOBO & Animal &\small{a colorful parrot, portrayed in vibrant fauvist colors, feathers bright and chattering away in a tropical canopy} \\ \hline
RTUB & SANVITO & Animal &\small{a playful dolphin, captured in watercolor blues, leaping joyfully above ocean waves, embodying freedom and grace} \\ \hline
fCgc & COOPBL & Animal &\small{a swift cheetah, rendered in dynamic cubist fragments, muscles tensed, darting across the plain in a blur of speed and agility} \\ \hline
INeC & COOPBL & Animal &\small{a serene swan, painted in impressionist pastels, gliding elegantly across a calm lake, its reflection a picture of tranquility} \\ \hline
VbTO & POSTINO & Animal &\small{a wise old elephant, captured in detailed charcoal, ambling through the jungle, its skin a tapestry of life's journeys} \\ \hline
dVDL & POPLAR & Food & \small{gingerbread} \\ \hline
sMjS & HOBO & Food & \small{sushi} \\ \hline
yoTb & POSTINO & Food & \small{pasta} \\ \hline
IRKd & COOPBL & Food & \small{donut} \\ \hline
LQln & POPLAR & Food & \small{croissant} \\ \hline
xCYU & HOBO & Food & \small{cookies} \\ \hline
AXOz & COOPBL & Food & \small{ice cream} \\ \hline
FXHh & POSTINO & Food & \small{orange} \\ \hline
quEP & COOPBL & Food & \small{juice splash} \\ \hline
gDwK & HOBO & Food & \small{toasted bread} \\ \hline
ZHex & SANVITO & Food & \small{smoked salmon atop a creamy dill spread on rye} \\ \hline
UzAB & SANVITO & Food & \small{iced matcha latte with a swirl of honey} \\ \hline
pGFZ & SANVITO & Food & \small{char-grilled oysters with a garlic butter sauce} \\ \hline
mRvP & COOPBL & Food & \small{warm pear tart tatin with a dollop of vanilla ice cream} \\ \hline
twhc & POPLAR & Food & \small{fresh mozzarella and tomato salad with basil pesto} \\ \hline
GeVa & HOBO & Food & \small{a refreshing elderflower spritz, effervescent and floral, combined with prosecco and a splash of soda water, adorned with a lemon twist for a light, celebratory drink} \\ \hline
krJi & SANVITO & Food & \small{a plate of fluffy pancakes, drizzled with maple syrup and topped with a handful of fresh blueberries} \\ \hline
TWyY & POPLAR & Food & \small{a charcuterie board, featuring thinly sliced prosciutto, salami, and a variety of aged cheeses} \\ \hline
fNQE & POSTINO & Food & \small{a vibrant summer salad, tossed with fresh greens, colorful edible flowers, and a light citrus vinaigrette} \\ \hline
CBJO & POSTINO & Food & \small{a berry-infused iced tea, sweetened just right and served with ice, garnished with fresh berries and a sprig of mint for a refreshing summer quencher} \\ \hline
fJTE & SANVITO & Material & \small{broken glass} \\ \hline
KdUz & HOBO & Material &\small{plastic wrap} \\ \hline
NMDh & COOPBL & Material &\small{marble granite} \\ \hline
ZrHl & POSTINO & Material &\small{metallic} \\ \hline
UmpZ & COOPBL & Material &\small{gold balloon} \\ \hline
hYnp & SANVITO & Material &\small{chainlink} \\ \hline
yDjP & POPLAR & Material & \small{watercolor} \\ \hline
sQBt & POSTINO & Material & \small{sequins} \\ \hline
aguy & HOBO & Material &\small{neon light} \\ \hline
LiXb & HOBO & Material &\small{colorful shaggy fur} \\ \hline
OLAe & SANVITO & Material &\small{purple paint brush stroke} \\ \hline
vBSJ & POPLAR & Material &\small{holographic dripping color} \\ \hline
oTYG & POSTINO & Material &\small{folk embroidered fabric} \\ \hline
XIqV & SANVITO & Material &\small{colorful christmas lights} \\ \hline
eCHs & HOBO & Material &\small{red and green holiday ornaments} \\ \hline
xkdr & POSTINO & Material &\small{sparkling frost crystals, covering the ground in a delicate, twinkling carpet} \\ \hline
RWIw & POPLAR & Material &\small{luminous moonstones, their surfaces alive with an ethereal, shifting glow} \\ \hline
cSFb & COOPBL & Material &\small{metallic dragonfly wings, catching light to reveal intricate, vibrant patterns} \\ \hline
almV & POPLAR & Material &\small{glistening fish scales, reflecting a rainbow of colors beneath clear waters,  depicted with vibrant impressionist strokes} \\ \hline
AMQN & COOPBL & Material &\small{glossy cherry wood, its surface smooth and reflecting a warm, deep sheen} \\
\end{tabular}  
}
\caption{  
\footnotesize{Full list of \ourbenchmark [English] benchmark. Part 1/2.}}
\label{tab:bench_en1}
\end{minipage}
\end{table} 

\begin{table}[htbp]
\vspace{-3mm}
\begin{minipage}[t]{1\linewidth}  
\centering  
\tablestyle{2pt}{1.1}  
\resizebox{1.0\linewidth}{!}  
{  
\begin{tabular}{l|l|l|>{\centering\arraybackslash}m{15cm}}  
Characters & Font Type & Category & Prompt \\  
\shline
TbtU & SANVITO & Nature &\small{fire} \\ \hline
AqWw & HOBO &Nature & \small{diftwood} \\ \hline
hmjG & POPLAR & Nature &\small{jungle vine} \\ \hline
aySt & COOPBL & Nature &\small{leafy pothos} \\ \hline
BeMc & HOBO & Nature &\small{lava} \\ \hline
XkFQ & SANVITO & Nature &\small{icicle} \\ \hline
CJZK & SANVITO & Nature &\small{decay} \\ \hline
hzRY & POSTINO & Nature &\small{pink flower petals} \\ \hline
npdN & POPLAR &Nature & \small{house plants} \\ \hline
xfEr & HOBO &Nature & \small{mossy rocks} \\ \hline
xFCA & HOBO &Nature & \small{lightning and rainclouds} \\ \hline
iNov & COOPBL &Nature & \small{botanical hand drawn illustration} \\ \hline
gLHs & POSTINO &Nature & \small{a dense canopy of rainforest trees} \\ \hline
nEOB & POPLAR &Nature & \small{snow-capped trees} \\ \hline
dIZl & POPLAR &Nature & \small{a field of wildflowers} \\ \hline
RfuY & SANVITO &Nature & \small{macro photography of dewdrops on a spiderweb, with morning sunlight creating rainbows} \\ \hline
VUkM & COOPBL &Nature & \small{a meadow bursting with wildflowers, their colors a vivid tapestry under the bright summer sun} \\ \hline
gJTD & COOPBL &Nature & \small{a coral reef, alive with color and bustling marine life, captured in the vivid colors of a fauvist painting} \\ \hline
GQei & POSTINO &Nature & \small{the northern lights dancing across the sky, a mesmerizing display of colors in the cold night air} \\ \hline
PqKD & POSTINO &Nature & \small{the majestic sight of a waterfall cascading down rocky terrains, enveloped in a misty spray, rendered in the rich, textured layers of an oil painting} \\ \hline
xyAa & POSTINO & Landscape& \small{harbor} \\ \hline
tlec & SANVITO & Landscape& \small{garden} \\ \hline
gmsN & HOBO & Landscape& \small{jungle} \\ \hline
kQUT & COOPBL & Landscape& \small{lighthouse} \\ \hline
dYQK & POSTINO & Landscape& \small{shopping mall} \\ \hline
JRkP & SANVITO & Landscape& \small{houses} \\ \hline
wzSO & HOBO & Landscape& \small{mountain} \\ \hline
GjLS & POSTINO & Landscape& \small{ruins} \\ \hline
zMVC & POPLAR & Landscape& \small{room} \\ \hline
rfib & SANVITO & Landscape& \small{temple} \\ \hline
FhoX & COOPBL & Landscape& \small{a cozy cottage with smoke} \\ \hline
DGnw & POPLAR & Landscape& \small{a narrow alleyway in an old city} \\ \hline
qAcs & POPLAR & Landscape& \small{a night sky ablaze with stars} \\ \hline
TvUH & POSTINO & Landscape& \small{volcano dripping with lava hyperrealistic} \\ \hline
IEMZ & COOPBL & Landscape& \small{a starlit sky above a quiet, sleeping village} \\ \hline
EpZC & HOBO & Landscape& \small{art nouveau painting of a female botanist surrounded by exotic plants in a greenhouse} \\ \hline
LrhR & COOPBL & Landscape& \small{snow-covered roottops glistened in the moonlight, with the streets below filled with the muffled sounds of footsteps and distant carolers} \\ \hline
IWBb & HOBO & Landscape& \small{cubist painting of a bustling city market with different perspectives of people and stalls} \\ \hline
FHYl & POPLAR & Landscape& \small{gothic painting of an ancient castle at night, with a full moon, gargoyles, and shadows} \\ \hline
deuO & SANVITO & Landscape& \small{a detailed drawing of a succulent garden, showcasing various textures and shades of green, with tiny flowers emerging} \\

\end{tabular}  
}
\caption{  
\footnotesize{Full list of \ourbenchmark [English] benchmark. Part 2/2.}}
\label{tab:bench_en2}
\end{minipage}
\end{table}

\begin{table}[htbp]
\begin{minipage}[t]{1\linewidth}  
\centering  
\tablestyle{2pt}{1.1}  
\resizebox{1.0\linewidth}{!}  
{  
\begin{tabular}{l|l|l|>{\centering\arraybackslash}m{15cm}}  
Characters & Font Type & Category & Prompt \\  
\shline
歌人水项 & SourceHanSansCNHeavy & Animal& \small{kitties} \\ \hline
水福木项 & SourceHanSansCNHeavy & Animal& \small{a butterfly flitting among wildflowers} \\ \hline
歌福走水 & SourceHanSansCNHeavy & Animal& \small{a wise old elephant, captured in detailed charcoal, ambling through the jungle, its skin a tapestry of life's journeys} \\ \hline
正福项人 & SourceHanSansCNHeavy & Food& \small{ice cream} \\ \hline
水正人歌 & SourceHanSansCNHeavy & Food& \small{smoked salmon atop a creamy dill spread on rye} \\ \hline
木项福沐 & SourceHanSansCNHeavy & Food& \small{a charcuterie board, featuring thinly sliced prosciutto, salami, and a variety of aged cheeses} \\ \hline
口正走沐 & SourceHanSansCNHeavy & Material& \small{broken glass} \\ \hline
歌走水项 & SourceHanSansCNHeavy & Material& \small{holographic dripping color} \\ \hline
走福正沐 & SourceHanSansCNHeavy & Material& \small{glistening fish scales, reflecting a rainbow of colors beneath clear waters,  depicted with vibrant impressionist strokes} \\ \hline
歌人沐口 & SourceHanSansCNHeavy & Nature& \small{lava} \\ \hline
歌口正人 & SourceHanSansCNHeavy & Nature& \small{lightning and rainclouds} \\ \hline
正木走口 & SourceHanSansCNHeavy & Nature& \small{the majestic sight of a waterfall cascading down rocky terrains, enveloped in a misty spray, rendered in the rich, textured layers of an oil painting} \\ \hline
沐水木口 & SourceHanSansCNHeavy & Landscape& \small{harbor} \\ \hline
口人木走 & SourceHanSansCNHeavy & Landscape& \small{a night sky ablaze with stars} \\ \hline
沐木福项 & SourceHanSansCNHeavy & Landscape& \small{cubist painting of a bustling city market with different perspectives of people and stalls} \\
\end{tabular}  
}
\caption{  
\footnotesize{Full list of \ourbenchmark [Chinese] benchmark.}}
\label{tab:bench_cn}
\end{minipage}
\begin{minipage}[t]{1\linewidth}  
\centering  
\tablestyle{2pt}{1.1}  
\resizebox{1.0\linewidth}{!}  
{  
\begin{tabular}{l|l|l|>{\centering\arraybackslash}m{15cm}}  
Characters & Font Type & Category & Prompt \\  
\shline
\begin{CJK}{UTF8}{min}い験か学\end{CJK} & SourceHanSansJPHeavy &Animal&  \small{firefly} \\ \hline
\begin{CJK}{UTF8}{min}いんあ学\end{CJK} &  SourceHanSansJPHeavy & Animal&\small{a spotted deer grazing in a meadow at dawn} \\ \hline
\begin{CJK}{UTF8}{min}あん赤十\end{CJK} & SourceHanSansJPHeavy & Animal& \small{a playful dolphin, captured in watercolor blues, leaping joyfully above ocean waves, embodying freedom and grace} \\ \hline
\begin{CJK}{UTF8}{min}日あ赤い\end{CJK} & SourceHanSansJPHeavy &  Food&\small{croissant} \\ \hline
\begin{CJK}{UTF8}{min}あさ日か\end{CJK} &  SourceHanSansJPHeavy & Food&\small{iced matcha latte with a swirl of honey} \\ \hline
\begin{CJK}{UTF8}{min}さ日か十\end{CJK} &  SourceHanSansJPHeavy & Food&\small{a refreshing elderflower spritz, effervescent and floral, combined with prosecco and a splash of soda water, adorned with a lemon twist for a light, celebratory drink} \\ \hline
\begin{CJK}{UTF8}{min}あ十かん\end{CJK} &  SourceHanSansJPHeavy &Material& \small{sequins} \\ \hline
\begin{CJK}{UTF8}{min}赤さ験十\end{CJK} & SourceHanSansJPHeavy &Material&  \small{purple paint brush stroke} \\ \hline
\begin{CJK}{UTF8}{min}学い赤十\end{CJK} &  SourceHanSansJPHeavy &Material& \small{glossy cherry wood, its surface smooth and reflecting a warm, deep sheen} \\ \hline
\begin{CJK}{UTF8}{min}あか験い\end{CJK}& SourceHanSansJPHeavy & Nature & \small{decay} \\  \hline
\begin{CJK}{UTF8}{min}験学日い\end{CJK} &SourceHanSansJPHeavy & Nature&  \small{snow-capped trees} \\ \hline
\begin{CJK}{UTF8}{min}験赤学さ\end{CJK} &  SourceHanSansJPHeavy &Nature& \small{a meadow bursting with wildflowers, their colors a vivid tapestry under the bright summer sun} \\ \hline
\begin{CJK}{UTF8}{min}かさん日\end{CJK} &  SourceHanSansJPHeavy&Landscape & \small{garden} \\ \hline
\begin{CJK}{UTF8}{min}ん学験十\end{CJK} & SourceHanSansJPHeavy & Landscape& \small{a starlit sky above a quiet, sleeping village} \\ \hline
\begin{CJK}{UTF8}{min}日赤さん\end{CJK} & SourceHanSansJPHeavy & Landscape& \small{a detailed drawing of a succulent garden, showcasing various textures and shades of green, with tiny flowers emerging} \\
\end{tabular}  
}
\caption{  
\footnotesize{Full list of \ourbenchmark [Japanese] benchmark.}}
\label{tab:bench_jp}
\end{minipage}
\begin{minipage}[t]{1\linewidth}  
\centering  
\tablestyle{2pt}{1.1}  
\resizebox{1.0\linewidth}{!}  
{  
\begin{tabular}{l|l|l|>{\centering\arraybackslash}m{15cm}}  
Characters & Font Type & Category & Prompt \\  
\shline

\begin{CJK}{UTF8}{mj}갑가맑달\end{CJK} &  SourceHanSansKRHeavy &Animal& \small{dragon} \\ \hline
\begin{CJK}{UTF8}{mj}강가달차\end{CJK} &  SourceHanSansKRHeavy &Animal& \small{a camel silhouetted against a desert sunset} \\ \hline
\begin{CJK}{UTF8}{mj}닭차갑나\end{CJK} &  SourceHanSansKRHeavy &Animal& \small{a colorful parrot, portrayed in vibrant fauvist colors, feathers bright and chattering away in a tropical canopy} \\ \hline
\begin{CJK}{UTF8}{mj}맑차갑가\end{CJK} &  SourceHanSansKRHeavy &Food& \small{juice splash} \\ \hline
\begin{CJK}{UTF8}{mj}달차바손\end{CJK} & SourceHanSansKRHeavy &  Food&\small{fresh mozzarella and tomato salad with basil pesto} \\ \hline
\begin{CJK}{UTF8}{mj}닭맑갑가\end{CJK} &  SourceHanSansKRHeavy &Food& \small{a berry-infused iced tea, sweetened just right and served with ice, garnished with fresh berries and a sprig of mint for a refreshing summer quencher} \\ \hline
\begin{CJK}{UTF8}{mj}맑손강달\end{CJK} &  SourceHanSansKRHeavy &Material& \small{marble granite} \\ \hline
\begin{CJK}{UTF8}{mj}강바손가\end{CJK} &  SourceHanSansKRHeavy & Material&\small{red and green holiday ornaments} \\ \hline
\begin{CJK}{UTF8}{mj}손달바강\end{CJK} &SourceHanSansKRHeavy & Material&  \small{luminous moonstones, their surfaces alive with an ethereal, shifting glow} \\ \hline
\begin{CJK}{UTF8}{mj}차닭갑나\end{CJK} &  SourceHanSansKRHeavy &Nature& \small{house plants} \\ \hline
\begin{CJK}{UTF8}{mj}바나닭차\end{CJK} &  SourceHanSansKRHeavy &Nature& \small{a field of wildflowers} \\ \hline
\begin{CJK}{UTF8}{mj}맑나달손\end{CJK} & SourceHanSansKRHeavy & Nature& \small{a coral reef, alive with color and bustling marine life, captured in the vivid colors of a fauvist painting} \\ \hline
\begin{CJK}{UTF8}{mj}나바갑강\end{CJK} &  SourceHanSansKRHeavy &Landscape& \small{room} \\ \hline
\begin{CJK}{UTF8}{mj}닭맑바나\end{CJK} & SourceHanSansKRHeavy & Landscape& \small{volcano dripping with lava hyperrealistic} \\ \hline
\begin{CJK}{UTF8}{mj}닭강가손\end{CJK} &  SourceHanSansKRHeavy &Landscape& \small{art nouveau painting of a female botanist surrounded by exotic plants in a greenhouse} \\

\end{tabular}  
}
\caption{  
\footnotesize{Full list of \ourbenchmark [Korean] benchmark.}}
\label{tab:bench_kr}
\end{minipage}
\end{table}

\begin{table}[htbp]
\begin{minipage}[t]{1\linewidth}  
\centering  
\tablestyle{2pt}{1.1}  
\resizebox{1.0\linewidth}{!}  
{  
\begin{tabular}{l|l|>{\centering\arraybackslash}m{15cm}}  
Characters & Font Type & Prompt \\  
\shline
AhSg & HOBO & \small{a red fox prowling through a snowy forest} \\ \hline
LwDK & POPLAR & \small{a butterfly flitting among wildflowers} \\ \hline
quEP & COOPBL & \small{juice splash} \\ \hline
krJi & SANVITO & \small{a plate of fluffy pancakes, drizzled with maple syrup and topped with a handful of fresh blueberries} \\ \hline
aySt & COOPBL & \small{leafy pothos} \\ \hline
PqKD & POSTINO & \small{the majestic sight of a waterfall cascading down rocky terrains, enveloped in a misty spray, rendered in the rich, textured layers of an oil painting} \\ \hline
\begin{CJK}{UTF8}{min}あか験い\end{CJK} & SourceHanSansJPHeavy & \small{decay} \\  \hline
木项福沐 & SourceHanSansCNHeavy & \small{a charcuterie board, featuring thinly sliced prosciutto, salami, and a variety of aged cheeses} \\
\end{tabular}  
}
\caption{  
\footnotesize{
Illustrating the font effect prompts presented in Figure~\ref{fig:Firefly} of the supplementary material are organized sequentially from the top left corner to the bottom right.}}
\label{tab:prompt_firefly}
\end{minipage}
\begin{minipage}[t]{1\linewidth}  
\centering  
\tablestyle{2pt}{1.1}  
\resizebox{1.0\linewidth}{!}  
{  
\begin{tabular}{l|l|>{\centering\arraybackslash}m{15cm}}  
Characters & Font Type & Prompt \\  
\shline

水福木项 & SourceHanSansCNHeavy & \small{a butterfly flitting among wildflowers} \\ \hline
正福项人 & SourceHanSansCNHeavy & \small{ice cream} \\ \hline
水正人歌 & SourceHanSansCNHeavy & \small{smoked salmon atop a creamy dill spread on rye} \\ \hline
口正走沐 & SourceHanSansCNHeavy & \small{broken glass} \\ \hline
歌走水项 & SourceHanSansCNHeavy & \small{holographic dripping color} \\ \hline
走福正沐 & SourceHanSansCNHeavy & \small{glistening fish scales, reflecting a rainbow of colors beneath clear waters,  depicted with vibrant impressionist strokes} \\ \hline
歌人沐口 & SourceHanSansCNHeavy & \small{lava} \\ \hline
口人木走 & SourceHanSansCNHeavy & \small{a night sky ablaze with stars} \\ \hline

\begin{CJK}{UTF8}{min}いんあ学\end{CJK} & SourceHanSansJPHeavy & \small{a spotted deer grazing in a meadow at dawn} \\ \hline
\begin{CJK}{UTF8}{min}日あ赤い\end{CJK} & SourceHanSansJPHeavy & \small{croissant} \\ \hline

\begin{CJK}{UTF8}{min}さ日か十\end{CJK} & SourceHanSansJPHeavy & \small{a refreshing elderflower spritz, effervescent and floral, combined with prosecco and a splash of soda water, adorned with a lemon twist for a light, celebratory drink} \\ \hline
\begin{CJK}{UTF8}{min}あ十かん\end{CJK} & SourceHanSansJPHeavy & \small{sequins} \\ \hline
\begin{CJK}{UTF8}{min}学い赤十\end{CJK} & SourceHanSansJPHeavy & \small{glossy cherry wood, its surface smooth and reflecting a warm, deep sheen} \\ \hline
\begin{CJK}{UTF8}{min}かさん日\end{CJK} & SourceHanSansJPHeavy & \small{garden} \\ \hline
\begin{CJK}{UTF8}{min}ん学験十\end{CJK} & SourceHanSansJPHeavy & \small{a starlit sky above a quiet, sleeping village} \\ \hline
\begin{CJK}{UTF8}{min}日赤さん\end{CJK} & SourceHanSansJPHeavy & \small{a detailed drawing of a succulent garden, showcasing various textures and shades of green, with tiny flowers emerging} \\ \hline

\begin{CJK}{UTF8}{mj}닭차갑나\end{CJK} & SourceHanSansKRHeavy & \small{a colorful parrot, portrayed in vibrant fauvist colors, feathers bright and chattering away in a tropical canopy} \\ \hline
\begin{CJK}{UTF8}{mj}맑차갑가\end{CJK} & SourceHanSansKRHeavy & \small{juice splash} \\ \hline
\begin{CJK}{UTF8}{mj}달차바손\end{CJK} & SourceHanSansKRHeavy & \small{fresh mozzarella and tomato salad with basil pesto} \\ \hline
\begin{CJK}{UTF8}{mj}닭맑갑가\end{CJK} & SourceHanSansKRHeavy & \small{a berry-infused iced tea, sweetened just right and served with ice, garnished with fresh berries and a sprig of mint for a refreshing summer quencher} \\ \hline
\begin{CJK}{UTF8}{mj}맑손강달\end{CJK} & SourceHanSansKRHeavy & \small{marble granite} \\ \hline
\begin{CJK}{UTF8}{mj}손달바강\end{CJK} & SourceHanSansKRHeavy & \small{luminous moonstones, their surfaces alive with an ethereal, shifting glow} \\ \hline
\begin{CJK}{UTF8}{mj}바나닭차\end{CJK} & SourceHanSansKRHeavy & \small{a field of wildflowers} \\ \hline
\begin{CJK}{UTF8}{mj}나바갑강\end{CJK} & SourceHanSansKRHeavy & \small{room} \\
\end{tabular}  
}
\caption{  
\footnotesize{Illustrating the font effect prompts presented in Figure~\ref{fig:Showcase_CJK} of the supplementary material are organized sequentially from the top left corner to the bottom right.}}
\label{tab:prompt_cjk}
\end{minipage}
\begin{minipage}[t]{1\linewidth}  
\centering  
\tablestyle{30pt}{1}  
\resizebox{0.6\linewidth}{!}  
{ 
\begin{tabular}{l|cc}
Category & CLIP↑ & CLIP-I↑\\ 
\shline
Nature &  29.16 & 84.94 \\
Material &  28.71 & 85.06 \\
Food & 30.70 & 85.35\\
Animal & 30.70 & 84.01 \\
Landscape & 27.91 & 83.78 \\
\hline
Overall & 29.44 & 84.63 \\
\end{tabular}
}
\caption{Categorical Quantitative Results of \ourname.}
\label{tab:category_result}
\vspace{-4mm}
\end{minipage}
\end{table}

\subsection{More Results for Chinese, Japanese and Korean Font}
\begin{figure}[t] 

\tiny
\centering
\vspace{2mm}

\includegraphics[width=0.975\textwidth]{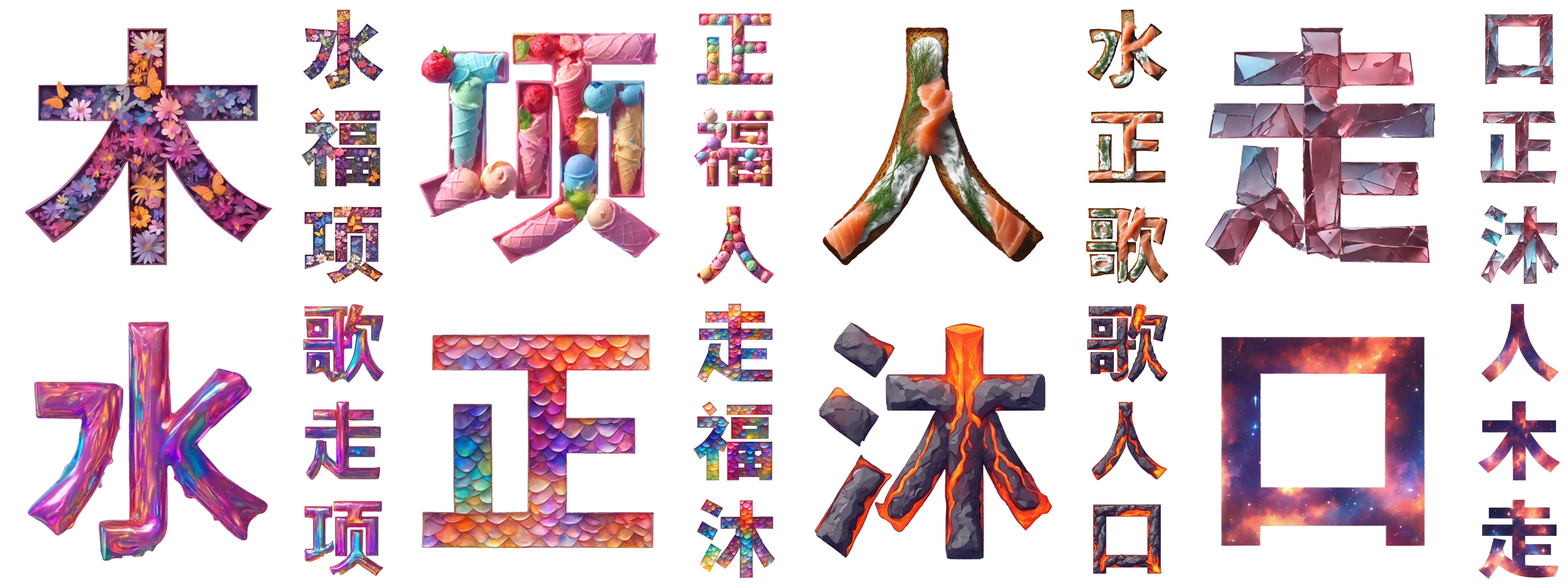}
\includegraphics[width=0.975\textwidth]{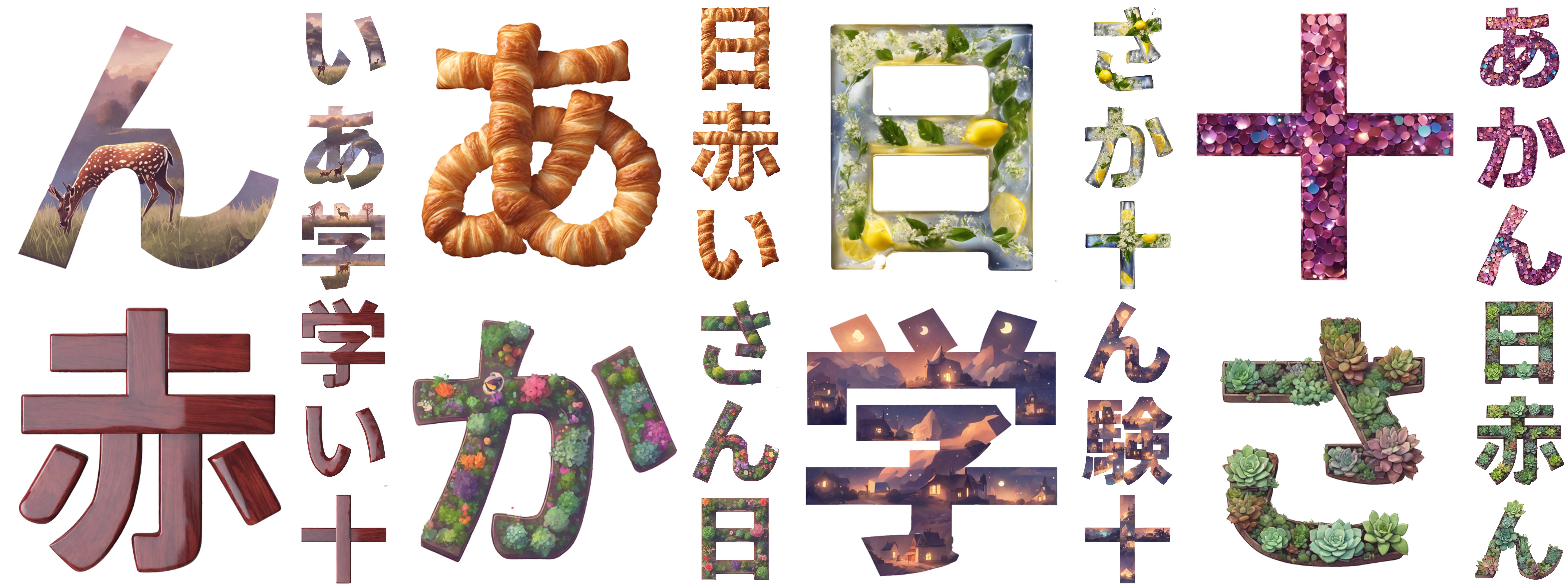}
\includegraphics[width=0.975\textwidth]{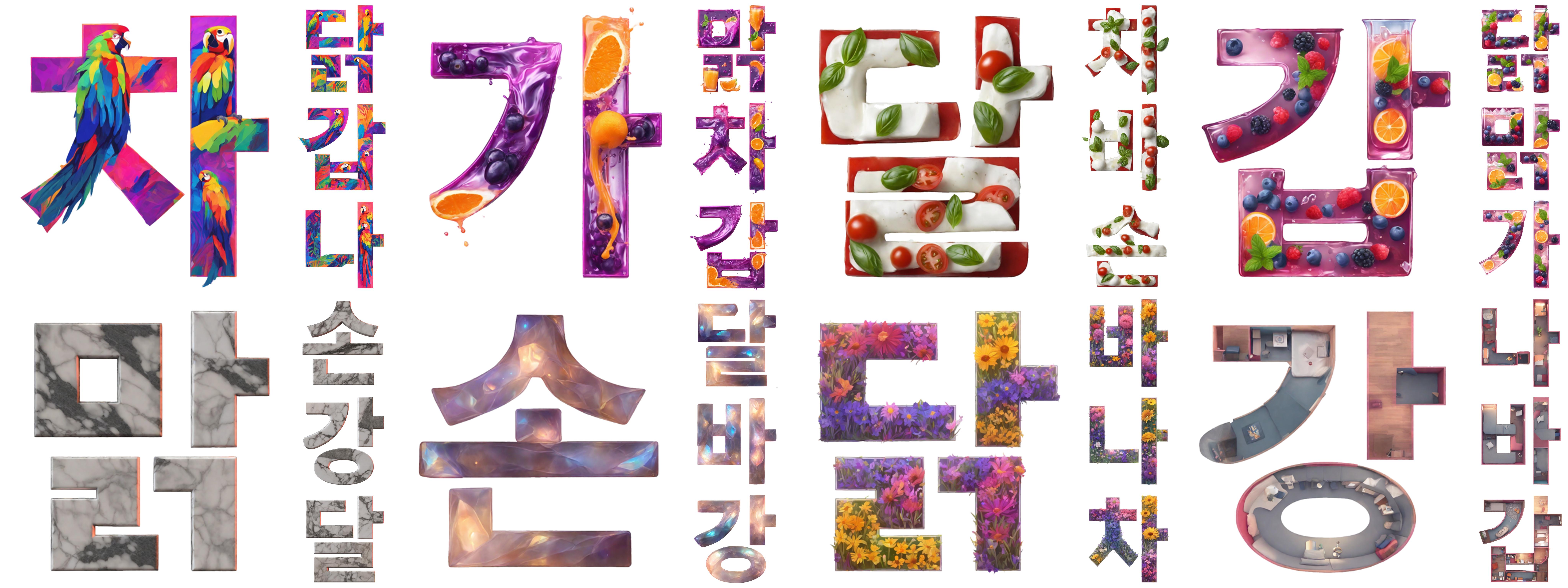}
\caption{\small More Qualitative Results for Chinese, Japanese and Korean.}
\label{fig:Showcase_CJK}
\vspace{-3mm}

\end{figure}

To illustrate \ourname's adeptness in generating content that features complex shapes, we present supplementary results for Chinese, Japanese, and Korean characters in Figure \ref{fig:Showcase_CJK}, with the corresponding prompts detailed in Table \ref{tab:prompt_cjk}.

\end{CJK*}

\subsection{More \ourname Results of Different Categories}
This section presents categorical results for \ourname, highlighting specific performance across different categories. According to Table \ref{tab:category_result}, Animal and Landscape emerge as particularly challenging categories for font effect generation.

\subsection{Discussion About the Choice of Reference Letter `R'.}
\begin{figure}
    \centering
    \includegraphics[width=0.9\textwidth]{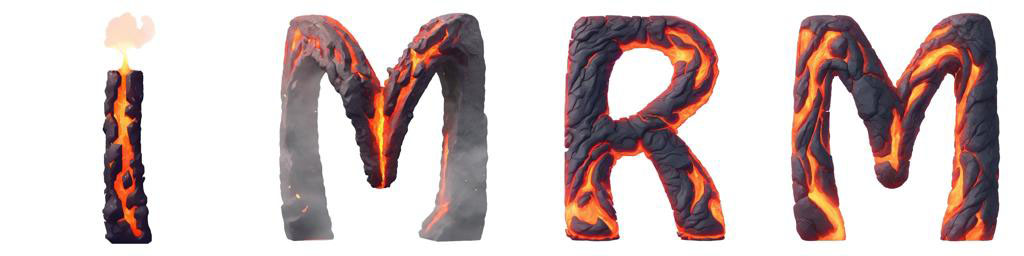}
    \caption{Shape-adaptive Effect Transfer results from different reference letter to target letter `M'. The left two images are the effect transfer results from reference letter `i' and the rest shows the effect transfer results from reference letter `R'.}
    \label{fig:ref_letter_choice}
\end{figure}

We provide more results to support our choice of reference character. As shown in Figure~\ref{fig:ref_letter_choice} and Table~\ref{tab:DINO}, using reference letter `R' is significantly better both qualitatively and quantitatively.

\begin{table}[]
\begin{minipage}[t]{1\linewidth}
\centering
{
\tablestyle{0.5pt}{1.2}
\begin{tabular}{l|c}
Method & DINO$\uparrow$\\
\shline
\ourname w. ref-letter `i'  & 60.79 \\
\ourname w. ref-letter `R'  & \textbf{67.07} \\ 
\end{tabular}
}
\caption{Font style consistency metric based on DINO with difference reference letter.}
\label{tab:DINO}
\end{minipage}
\end{table}

\subsection{Discussion About Font Shape Readability and Text-effect Strength.}
Shape-adaptive refinement model allows flexible control between readability and text-effect strength via noise strength. The default noise strength is 0.8. This setting allows the model to  alter the font shape. However, there are instances where the model opts to preserve the original shape.

We empirically find that this phenomenons occurs primarily in two scenarios: 1) When the user's prompt includes content like metal or marble, which lack a regular shape pattern, leading the model to retain the original font shape. See "\begin{CJK}{UTF8}{mj}맑손강달\end{CJK}" (marble granite)  in Figure~\ref{fig:MoreFirefly}. 2) When the prompt involves a scene with a background, similar to some of the cropped scenes used in our training data (refer to the right two columns of Figure~\ref{fig:dalle_training}). The border of a scene's background can be challenging to refine due to its potential vastness, so the model tends to preserve the original canvas border based on its learning from the data. For example, in the case of the characters "\begin{CJK*}{UTF8}{gbsn}口人木走\end{CJK*}" (a night sky ablaze with stars) in Figure~\ref{fig:MoreFirefly}, the model treats it as a scene with a background and maintains its original shape. However, for "\begin{CJK}{UTF8}{min}んいあ学\end{CJK}" (a spotted deer grazing in a meadow at dawn) in Figure~\ref{fig:MoreFirefly}, which involves completing complex elements like missing deer legs, the task exceeds the model's refinement capability. Hence, the model treats the deer as part of the background and preserves the original text shape.

\end{document}